\documentclass[10pt,twocolumn,letterpaper]{article}

\usepackage{iccv}              

\definecolor{iccvblue}{rgb}{0.21,0.49,0.74}
\usepackage[pagebackref,breaklinks,colorlinks,allcolors=iccvblue]{hyperref}

\def\paperID{8511} 
\def\confName{ICCV}
\def\confYear{2025}

\title{SweetTok: Semantic-Aware Spatial-Temporal Tokenizer for Compact Video Discretization}

\author{Zhentao Tan$^\ast$, Ben Xue$^\ast$, Jian Jia, Junhao Wang, Wencai Ye, Shaoyun Shi, Mingjie Sun\\ Wenjin Wu, Quan Chen$^\dagger$, Peng Jiang\\
 Kuaishou Technology, Beijing, China\\
{\tt\small \{tanzhentao03, xueben, jiajian, wangjunhao05, yewencai, shishaoyun, } \\
{\tt\small sunmingjie, wuwenjin, chenquan06, jiangpeng\}@kuaishou.com}}

\begin{document}
\maketitle
\begin{abstract}

This paper presents the \textbf{S}emantic-a\textbf{W}ar\textbf{E} spatial-t\textbf{E}mporal \textbf{T}okenizer (SweetTok), a novel video tokenizer to overcome the limitations in current video tokenization methods for compacted yet effective discretization. Unlike previous approaches that process flattened local visual patches via direct discretization or adaptive query tokenization, SweetTok proposes a decoupling framework, compressing visual inputs through distinct spatial and temporal queries via  \textbf{D}ecoupled \textbf{Q}uery \textbf{A}uto\textbf{E}ncoder (DQAE). 
This design allows SweetTok to efficiently compress video token count while achieving superior fidelity by capturing essential information across spatial and temporal dimensions. Furthermore, we design a \textbf{M}otion-enhanced \textbf{L}anguage \textbf{C}odebook (MLC) tailored for spatial and temporal compression to address the differences in semantic representation between appearance and motion information. SweetTok significantly improves video reconstruction results by \textbf{42.8\%} w.r.t rFVD on UCF-101 dataset.
With a better token compression strategy, it also boosts downstream video generation results by \textbf{15.1\%} w.r.t gFVD. Additionally, the compressed decoupled tokens are imbued with semantic information, enabling few-shot recognition capabilities powered by LLMs in downstream applications.

\end{abstract}    
\section{Introduction}
\label{sec:intro}


\begin{figure}[t]
\centering
\includegraphics[width=0.9\linewidth]{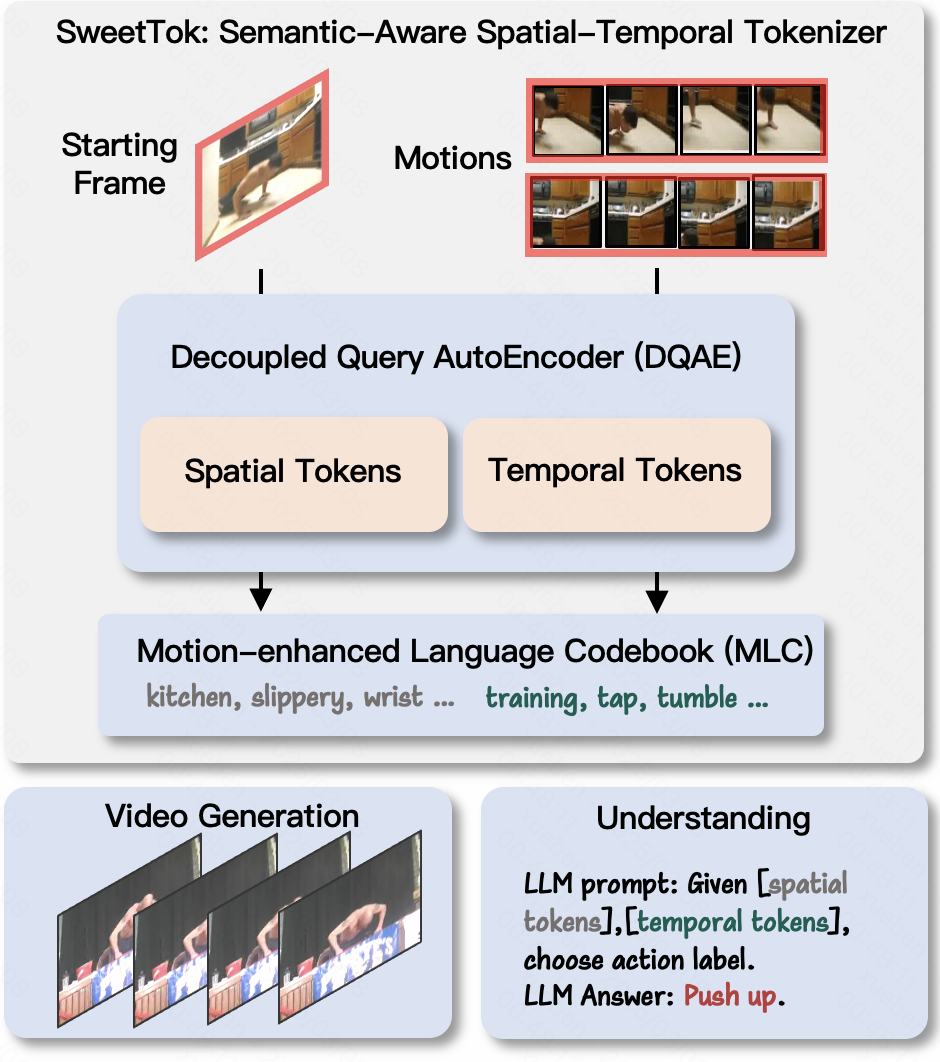}
\caption{
Illustration of our framework.
We build a compact visual latent space by reducing token count in a decoupled style and leveraging motion-enhanced semantic text embedding.
The encoded tokens can be applied to downstream tasks, such as generation and understanding.
}
\label{fig:title_fig}
\vspace{-5mm}
\end{figure}

Visual tokenizers \cite{van2017neural, chang2022maskgit, wang2024omnitokenizer, villegas2023cvivit, ge2022tats, yu2023magvit, yu2023magvit-v2, wang2024larp} are emerging as essential components in the field of modern computer vision models, particularly in the generation \cite{ge2022tats, yu2023magvit-v2, yoo2023towards} and understanding \cite{wang2024omnivid, jin2024videolavit, sun2023emu, lin2023video-llava, wang2024internvideo2} of vision data. These tools convert visual inputs into discrete tokens, capturing essential temporal and spatial features that facilitate advanced analysis by formulating visual-related tasks as a token prediction process.

Compression ratio and reconstruction fidelity are vital criteria for evaluating a tokenizer. Recent visual tokenizers, especially video tokenizers \cite{wang2024omnitokenizer, villegas2023cvivit, yu2023magvit-v2} typically retain a low compression ratio.
This is because visual tokens are usually derived from 2D patches \cite{dosovitskiy2020image_16x16} or 3D tubes \cite{wang2024omnitokenizer, ge2022tats} which preserve location relationships (e.g., each token corresponds to a specific region of input \cite{titok}), 
leading to redundancy in both spatial and temporal dimensions. Most recent work LARP \cite{wang2024larp} quantizes the flattened video patches through adaptive holistic queries to achieve high compression ratio. However, it is observed that directly flattening video tokens into sequence may lead to difficulty in learning intertwined spatial temporal information resulting in low reconstruction performance. 
Therefore, a new compression method needs to be proposed, one that takes into account the spatiotemporal properties of video.

Another issue, meanwhile, is that a higher compression ratio typically results in a greater loss of reconstruction details.
To complement visual information during compression, one common strategy is to introduce pretrained language embeddings as the latent codebook~\cite{liu2024lqae, yu2024spae, zhang2024vqct} ,
 leveraging their semantic representation capabilities.
However, previous works primarily focus on image modality, overlooking the relationships between text and motion in video domain.

To address existing limitations, we propose SweetTok -- \textbf{S}emantic-a\textbf{W}ar\textbf{E} spatial-t\textbf{E}mporal \textbf{T}okenizer -- as illustrated in Figure \ref{fig:title_fig}.  
Considering the heterogeneous redundancy in static images and dynamic frames, we propose the \textbf{D}ecoupled \textbf{Q}uery \textbf{A}uto\textbf{E}ncoder (DQAE) to compress spatial and temporal information into separate learnable queries. 
Different from previous works \cite{titok, blip2, wang2024larp}, our findings indicate that coupling the compression of spatiotemporal information increases the difficulty for the decoder to learn the motion information of the same pixel across consecutive frames. 
Thus, taking the decoupled spatial and temporal queries as inputs, we devise a strategy of spatial decoding followed by temporal decoding to achieve a separate reconstruction of the spatial and temporal dimensions of visual information. Additionally, the decoupled spatiotemporal reconstruction approach naturally allows for finetuning on image data, making our SweetTok flexible to image reconstruction task. 

Furthermore, to integrate the semantic information inherent in pre-traiend language model, we design a \textbf{M}otion-enhanced \textbf{L}anguage \textbf{C}odebook (MLC) tailored for spatial and temporal compression addressing the differences in semantic representation between spatial and temporal information.
Specifically, we design two language-based codebooks based on the part of speech, using nouns and adjectives for spatial static information and verbs and adverbs for temporal motion information. By incorporating language-based codebooks, the learnable compressed queries can also be easily adapted to downstream visual understanding tasks by in-context learning of LLM.

Exhaustive experiments demonstrate the effectiveness of SweetTok. Compared with vanilla video tokenizer without token compression (OmniTok\cite{wang2024omnitokenizer}), SweetTok improves rFVD by 52.3\% on UCF-101 \cite{soomro2012ucf101} using only 25\% of the tokens. 
Compared with vanilla query-based tokenizer (LARP \cite{wang2024larp}), SweetTok reduces rFVD from 35.15 to 20.46 and gFVD from 99 to 84, on UCF-101. 
By directly finetuning the decoupled spatial branch on the ImageNet-1k \cite{deng2009imagenet}, SweetTok also demonstrates a substantial improvement in rFID, decreasing it from 0.59 to 0.37.



In summary, our work makes the following key contributions:

\begin{itemize}
    \item 
We introduce SweetTok, a cutting-edge video tokenizer that achieves the state-of-the-art reconstruction fidelity with a high compression ratio via spatial-temporal decoupling and decoupled query autoencoder (DQAE), reaching a ``\textit{sweet spot}'' between compression and fidelity. 
    \item 
We propose a motion-enhanced language codebook (MLC) to more effectively capture the action information embedded in the video modality, thereby improving reconstruction quality and supporting downstream video understanding tasks.
    \item 
We perform extensive experiments to verify the effectiveness of SweetTok, which exhibits the state-of-the-art performance on video reconstruction, image reconstruction, and class-conditional video generation tasks, leading by a large margin of \textbf{42.8\%}, \textbf{37.2\%}, and \textbf{15.1\%}.
\end{itemize}


\begin{figure*}[t]
\centering
\includegraphics[width=\linewidth]{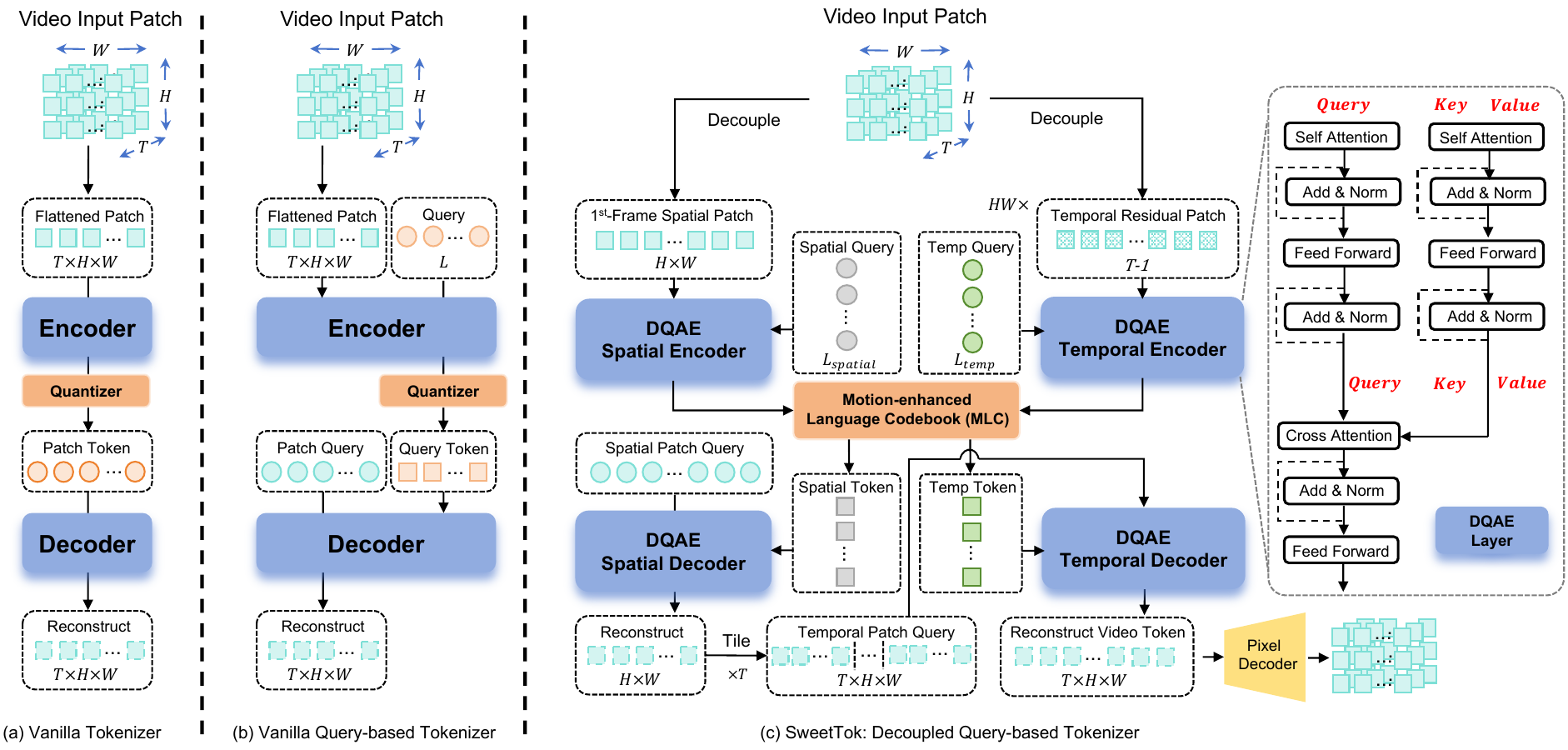}
\caption{Pipeline overview. \textbf{(a) Vanilla video tokenizers} directly quantize flattened video patches. \textbf{(b) Vanilla query-based tokenizers} compress flattend video patches into adaptive queries. \textbf{(c) SweetTok} proposes \textbf{decoupled query-based autoencoder} (\textbf{DQAE}, \S\ref{sec:3.1}). The spatial encoder quantizes the first frame's patch embeddings, while the temporal encoder quantizes residual between consecutive frames. The spatial decoder reconstructs the first frame's patches, replicates them $T$ times, and passes them to the temporal decoder for final information fusion and reconstruction. 
It also proposes \textbf{motion-enhanced language codebook} (\textbf{MLC}, \S\ref{sec:3.2}) to complement reconstructed video information via action-related language semantics.
}
\label{fig:main_fig_2}
\end{figure*}

\section{Background}

\subsection{Visual Tokenizer With Vector Quantization}




Exploring visual tokenizers and their applications in generative models has led to significant advancements in image/video-related tasks. The general idea is to discretize visual data into tokens, then tasks like visual generation \cite{chang2022maskgit, yu2021vim, yu2022scaling, ge2022tats} \& understanding \cite{dosovitskiy2020image_16x16, carion2020detr, wang2024omnivid, jin2024lavit, jin2024videolavit, sun2023emu, lin2023video-llava, wang2024internvideo2} can be tackled in a sequence prediction style as natural language processing \cite{devlin2018bert, radford2019language, touvron2023llama}. 
Our work belongs to the series of Vector Quantized Variational AutoEncoder (VQ-VAE) \cite{van2017neural, razavi2019vqvae-v2} tokenizers, which introduce a discrete latent space for continuous VAE \cite{kingma2013vae} encoder-decoder structure. It typically encodes a high-dimensional image into a low-dimensional latent representation, then queries the nearest index from a learnable codebook to quantize the latent vector, and finally decodes back reversely to reconstruct the raw input signal.
Since this type of tokenizer acquires reconstruction loss, it can maintain high-level semantic and low-level details of input vision.
VQGAN \cite{esser2020taming} adopted adversarial training loss to improve high-frequency details. ViT-VQGAN \cite{yu2021vim} upgraded encoder-decoder with vision-transformer (ViT) architecture \cite{dosovitskiy2020image_16x16} and further boosted results.
TiTok \cite{titok} replaced 2D image structure with 1D sequence latent representation, then used a self-attention transformer \cite{vaswani2017attention} to compress token number. 

However, the above methods can only process image data. For video modality, TATS \cite{ge2022tats} used 3D-CNN to encode video patches and adopted sliding windows to deal with long-term relations. CViViT \cite{villegas2023cvivit} used ViT \cite{dosovitskiy2020image_16x16} structure to encode spatial patches and then adopted a causal transformer to model temporal information. OmniTokenizer \cite{wang2024omnitokenizer} and MAGVIT \cite{yu2023magvit, yu2023magvit-v2} adopted similar transformer architecture and introduced image pre-training to improve video tokenizer. LARP \cite{wang2024larp}, on the other hand, introduces to compress flattened video patches into adaptive holistic queries with the guidance of a pre-trained auto-regressive model.
In this paper, we inherit the popular spatial-temporal decomposition design for video data. 

\subsection{Language-based Latent Codebook} 


The codebooks learned by vanilla VQ-VAEs are not interpretable with lexical meanings. Therefore, many works attempt to utilize pretrained language models embedding codebooks to enhance semantics. LQAE \cite{liu2024lqae} replaced the visual codebook with frozen word embeddings from BERT \cite{devlin2018bert}. 
SPAE \cite{yu2024spae} quantized image latent space in a pyramid structure to preserve semantic information from low-level to high-level. It also used large language model (LLM) codebook \cite{chowdhery2023palm} so that the encoded image token can be directly adapted to visual understanding tasks through in-context learning \cite{brown2020gpt3} ability of LLM. We follow this evaluation pipeline for few-shot classification in our paper.
V2L-Tokenizer \cite{zhu2024v2l-tokenizer} utilized CLIP \cite{radford2021clip} pretrained encoder and injected a learnable projector to align visual-text latent space implicitly. 
VQCT \cite{zhang2024vqct} replaced the projector with graph convolution networks \cite{kipf2016semi} to consider the relationship between vocabularies. 
Furthermore, De-Diffusion \cite{wei2024de-diffusion} directly encoded image into plain text as latent space interface and decodes back through a text-to-image (T2I) diffusion model \cite{rombach2021sd}.
However, these studies primarily focus on the image modality.
In this paper, we explore the design of the language codebook specifically for the video modality by splitting the codebook according to the video's spatial-temporal attribute.

\label{sec:background}

\section{Method}

\subsection{Preliminary}
\label{sec:prelim}

A typical visual vector-quantization (VQ) model \cite{ge2022tats, wang2024omnitokenizer, yu2023magvit, yu2023magvit-v2} contains three parts: encoder $\mathcal{E}$, decoder $\mathcal{D}$ and latent quantizer $\mathcal{Q}$ as shown in Figure \ref{fig:main_fig_2} (a). Take video modality as an example, given a video input $x \in \mathbb{R}^{T\times H\times W\times3}$, where $T$ represents the temporal length and $H\times W$ denotes spatial resolution, encoder $\mathcal{E}(x)$ projects it into latent space $\mathbb{Z}\in \mathbb{R}^{N \times D}$, where $D$ is latent dimension and $N$ is token number, 
A quantizer $\mathcal{Q}$ is constructed in this latent space $\mathbb{Z}$ by querying the nearest neighbor in codebook $C \in \mathbb{R}^{L_c\times D}$, where $L_c$ is codebook size. 
Then $\mathcal{D}$ decodes latent space back to pixel space and applies self-supervised reconstruction loss:
\begin{equation}
\mathcal{L}_{rec}(x, \mathcal{D}(\mathcal{Q}(\mathcal{E}(x)))).
\end{equation}
\vspace{-5mm}

Unlike traditional visual VQ models, LARP \cite{wang2024larp} quantizes flattened video patches into holistic queries $\mathbf{Q} \in \mathbb{R}^{L \times D}$ as shown in Figure \ref{fig:main_fig_2} (b), where $L$ is the adaptive query size. As shown in Equation (2), the encoder $\mathcal{E}$ processes concatenated flattened video patches $\mathbf{E} = flatten(x) \in \mathbb{R}^{N \times D}$ and query tokens $\mathbf{Q} \in \mathbb{R}^{L \times D}$, outputting $\mathbf{Z_Q} \in \mathbb{R}^{L \times D}$ for quantization. The decoder $\mathcal{D}$ reconstructs the video patches $\tilde{\mathbf{x}} \in \mathbb{R}^{T\times H\times W\times3}$ from the concatenation of learnable video queries $\mathbf{E_Q} \in \mathbb{R}^{N \times D}$ and the query discrete embeddings $\tilde{\mathbf{Z}}_\mathbf{Q}$.

\vspace{-4mm}
\begin{equation}
\mathbf{Z_E}||\mathbf{Z_Q}=\mathcal{E}(\mathbf{E} || \mathbf{Q}), \tilde{\mathbf{Z}}_\mathbf{Q} = \mathcal{Q}(\mathbf{Z_Q}), \tilde{x} = \mathcal{D}(\mathbf{E_Q} || \tilde{\mathbf{Z}}_\mathbf{Q}).
\end{equation}
\vspace{-5mm}

However, directly compressing video from flattened patches is challenging due to intertwined redundant temporal data and low-level spatial content, leading to sub-optimal performance. Thus, we designed SweetTok to balance reconstruction performance with a high compression ratio. Details are elaborated in Section \ref{sec:3.2_main}.


\subsection{Decoupled Spatial-Temporal Tokenization}
\label{sec:3.2_main}

As noted in Section \ref{sec:prelim}, directly quantizing video data via flattened patches hampers model learning due to intertwined redundant temporal and complex spatial information. Thus, we propose separately quantizing spatial and temporal dimensions before combining them to reconstruct the video, following the divide-and-conquer principle. The decoupling strategy enables high compression while ensuring higher fidelity. The main pipeline is shown in Figure \ref{fig:main_fig_2} (c).

\subsubsection{Patchify}

Given a video frame sequence $x \in \mathbb{R}^{T\times H\times W\times3}$, we select the first frame $x_1$ as a reference for spatial information,  the remaining $T-1$ frames $x_{2:T}$ for temporal information, following the strategy in \cite{wang2024omnitokenizer}. 
We apply two patch kernels $\mathcal{P}_s, \mathcal{P}_t$ with shapes $p_h \times p_w$ and $p_t \times p_h \times p_w$ to $x_1$ and $x_{2:T}$ separately, generating $v_s \in \mathbb{R}^{1 \times \frac{H}{p_h} \times \frac{W}{p_w}\times D}$ and $v_t \in \mathbb{R}^{\frac{T-1}{p_t} \times \frac{H}{p_h} \times \frac{W}{p_w} \times D}$ shown below:
\begin{equation}
    v_s = \mathcal{P}_s (x_1), v_t = \mathcal{P}_t (x_{2:T})
\end{equation}

$v_s$ and $v_t$ are inputs for transformer-based autoencoder, where $v_s$ contains spatial information, $v_t$ contains temporal information. 
In practice, for a video with 17 frames and a resolution of $256 \times 256$, we set $(p_t,p_h,p_w)$ to (4, 8, 8), thus patchify frames into $v_s$ with shape $1 \times 32 \times 32$ and $v_t$ with shape $4 \times 32 \times 32$. We use $t=\frac{T-1}{p_t}=4$ to denote $v_t$'s length.


\subsubsection{Decoupled Query AutoEncoder (DQAE)}
\label{sec:3.1}

To decouple spatial and temporal dimensions, we need to compress video patches separately along each dimension. Inspired by \cite{titok, wang2024larp} and Q-Former \cite{blip2}, we compress each dimension into an adaptive query tokens via cross-attention interactions. As an innovation, we recursively inject these cross-attention query modules into transformer-based autoencoder to transfer information, forming our DQAE module shown in the right gray box of Figure \ref{fig:main_fig_2} (c). For the rest of our paper, we use $\mathcal{E}_{DQAE}$ and $\mathcal{D}_{DQAE}$ as the encoder and decoder of the $DQAE$ module for simplicity.

\paragraph{Spatial Tokenization.} We observe that for most video parts, the first frame holds the most spatial information, so we use $v_s$ as the input to $\mathcal{E}_{DQAE_s}$ for quantization shown below:

\vspace{-4mm}
\begin{equation}
    \mathbf{Z}_{\mathbf{Q_s}} = \mathcal{E}_{DQAE_s} (\mathbf{Q_s}, v_s), \tilde{\mathbf{Z}}_{\mathbf{Q_s}} = \mathcal{Q}_{MLC} (\mathbf{Z}_{\mathbf{Q_s}}),
\end{equation}

where $\mathbf{Q_s} \in \mathbb{R}^{L_{spatial} \times D}$ are the learnable spatial query embeddings and $\mathbf{Z_\mathbf{Q_s}} \in \mathbb{R}^{L_{spatial} \times D}$ are the output embeddings encoding information from the first frame patches $v_s$. $\tilde{\mathbf{Z}}_{\mathbf{Q_s}} \in \mathbb{R}^{L_{spatial} \times D}$ are the quantized spatial token embeddings and $\mathcal{Q}_{MLC}$ stands for Motion-enhanced Language Codebook (MLC) quantizer which will be elaborated in the following section. After quantization, we inject our informative spatial token embeddings $\tilde{\mathbf{Z}}_{\mathbf{Q_s}}$ into a learnable spatial patch queries $\mathbf{Q}_{v_s}$ through decoder $\mathcal{D}_{DQAE_s}$ as below:

\vspace{-2mm}
\begin{equation}
    \tilde{v}_s = \mathcal{D}_{DQAE_s}(\mathbf{Q}_{v_s}, \tilde{\mathbf{Z}}_{\mathbf{Q_s}}),
\end{equation}

where $\tilde{v}_s$ are the reconstructed first frame video patches. The temporal component which will be stated in the following, combined with $\tilde{v}_s$ is used for the final video reconstruction. We set $L_{spatial} =256$ in real implementation.

\paragraph{Temporal Tokenization.} It is observed that for video data, there is much redundancy along temporal dimension. It motivates us to employ frame-wise residual $\Delta v_t = (\Delta v_t^1, \Delta v_t^2, ..., \Delta v_t^k)_{k = \frac{H}{p_h} \times \frac{W}{p_w}}$, where $\Delta v_t^i = v_s^i - v_t^i$, for tokenization. We use the first frame of the video for frame-wise residual because the spatial tokenization phase reconsturct it. Then, the residual $\Delta v = (\Delta v_1, \Delta v_2, ..., \Delta v_t)_{t = \frac{T-1}{p_t}}$ is input to $\mathcal{E}_{DQAE_t}$ for temporal compression, as shown below:

\vspace{-4mm}
\begin{equation}
\mathbf{Z}_{\mathbf{Q_t}} = \mathcal{E}_{DQAE_t} (\mathbf{Q_t}, \Delta v), \tilde{\mathbf{Z}}_{\mathbf{Q_t}} = \mathcal{Q}_{MLC} (\mathbf{Z}_{\mathbf{Q_t}}), 
\end{equation}

where $\mathbf{Q_t} \in \mathbb{R}^{L_{temporal} \times D}$ are the learnable temporal query embeddings and $\mathbf{Z_\mathbf{Q_t}} \in \mathbb{R}^{L_{temporal} \times D}$ are the output embeddings encoding information of the frame-wise residual. $\tilde{\mathbf{Z}}_{\mathbf{Q_t}} \in \mathbb{R}^{L_{temporal} \times D}$ are the quantized frame-wise temporal residual token embeddings. In practical implementation, $L_{temporal} = 1024$. After reconstructing the first frame patches, we recover the entire video patches by combining $\tilde{v}_s$ with the frame-wise quantized residual $\tilde{\mathbf{Z}}_{\mathbf{Q_t}}$. We tiled $\tilde{v}_s$ for $t$ times and sent it to $\mathcal{D}_{DQAE_t}$ shown below:

\vspace{-2mm}
\begin{equation}
    \tilde{v} = \mathcal{D}_{DQAE_t} ([\tilde{v}_s|| \cdots || \tilde{v}_s], \tilde{\mathbf{Z}}_{\mathbf{Q_t}}),
\end{equation}

where $\tilde{v}$ is the reconstructed video patches. The vectors $\tilde{v}$ reside in the latent space, necessitating the use of a ``pixel decoder" $\mathcal{D}_{pixel}$ to reconstruct the video data, as illustrated below:

\begin{equation}
    \tilde{x} = \mathcal{D}_{pixel} (\tilde{v}).
\end{equation}

The whole $DQAE$ is supervised by the reconstruction loss $\mathcal{L}_{rec}$ containing a $L_2$ loss, a LPIPS perception loss: $\mathcal{L}_{Lpips}$, a quantizer loss: $\mathcal{L}_{vq}$ and a GAN loss: $\mathcal{L}_{g}$ following the principle \cite{esser2020taming}. 

\subsubsection{Motion-enhanced Language Codebook (MLC)}
\label{sec:3.2}

To mitigate information loss during compression, we introduce an language codebook (LC) quantizer to enhance semantic richness. 
The Previous works~\cite{liu2024lqae, yu2024spae} have shown that text representations can enhance image VQ-VAEs , as the text provides additional semantic information from pre-trained language models. 
However, previous works mainly focus on the relationship between static image appearance and text semantics~\cite{zhang2024vqct}.
Our experiment in Table \ref{tab:alblation arch ucf} shows is insufficient for video data. 

\begin{figure}[t]
\centering
\includegraphics[width=\linewidth]{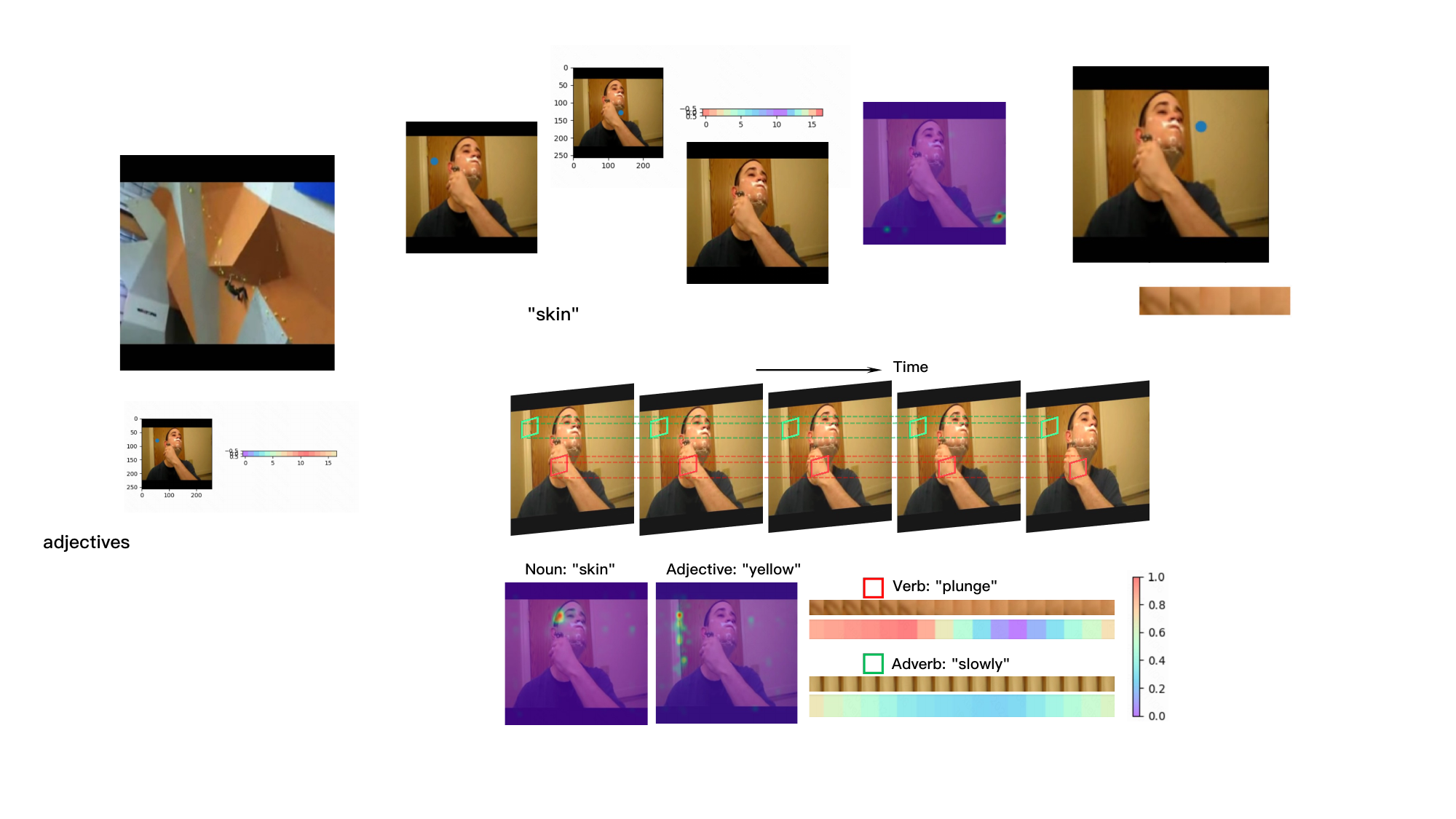}
\caption{
The semantics of spatial-temporal ``words''. The attention weights of the last encoder's cross-attention layer are visualized via heatmap, showing the visual regions corresponding to the related latent words. 
}
\vspace{-3mm}
\label{fig:semantic_align}
\end{figure}

To address this, we propose a Motion-enhanced Language Codebook (MLC), where the video motion information is enhanced via action-related vocabularies. 
Specifically, we split the dictionary into four subsets: nouns, adjectives, verbs, and adverbs.
Intuitively, static and appearance information is typically embedded in nouns and adjectives, while motion information is generally embedded in verbs and adverbs.
Therefore, we choose nouns and adjectives for spatial query tokens $\mathbf{Q_s}$, and verbs and adverbs for temporal query tokens $\mathbf{Q_t}$.
Figure \ref{fig:semantic_align} also shows that the encoded latent words by SweetTok capture semantic meanings related to both visual appearance and motion.

As for details, we first extract candidate vocabularies of the whole dataset from video captions. 
Afterward, we extract CLIP~\cite{radford2021clip} text embedding of these vocabularies to fill in the columns of our codebook $C \in \mathbb{R}^{L\times D}$.
We utilize a graph convolution network $\mathcal{F}$ to project CLIP embeddings \cite{radford2021clip} into the visual latent space. Graph edges are constructed when a pair of ``spatial-spatial'', ``spatial-temporal'' or ``temporal-temporal'' words co-occur within a 5-token window in the current video caption.

Given two encoded continuous latent vectors: $z_s \in \mathbf{Q_s}$ and $z_t \in \mathbf{Q_t}$, $z_s$ is passed through spatial quantization codebook, and $z_t$ is passed through temporal quantization codebook. The quantized $\hat{z}_s$ and $\hat{z}_t$ are obtained by nearest neighbor searching:
\begin{gather}
z_s, z_t = \mathcal{E}(x), \\
\hat{z}_s = \mathcal{F}(c_i), i = \mathop{\arg\min}\limits_{c_i \in C_{noun} \cup C_{adj}} || z_s - \mathcal{F}(c_i) ||, \\
\hat{z}_t = \mathcal{F}(c_i), i = \mathop{\arg\min}\limits_{c_i \in C_{verb} \cup C_{adv}} || z_t - \mathcal{F}(c_i) ||.
\end{gather}

Finally, the gradient is passed to the encoder via vector-quantization commitment loss 
 proposed in \cite{van2017neural}, 
 a common method to approximate differentiability ($sg[\cdot]$ stands for stop-gradient operator):
\begin{gather}
\label{eq:vq_loss}
\mathcal{L}_{vq} = || sg[z_s] - \mathcal{Q}(z_s)||^2 + || z_s - sg[\mathcal{Q}(z_s)] ||^2  \\ \notag
+ || sg[z_t] - \mathcal{Q}(z_t)||^2 + || z_t - sg[\mathcal{Q}(z_t)] ||^2 
\end{gather}

\label{sec:method}

\section{Experiments}
\subsection{Experiments Settings}

\paragraph{Dataset.} 
We evaluate the tokenization performance of SweetTok on video datasets, including UCF-101 \cite{soomro2012ucf101}, and Kinetics-600 \cite{kay2017kinetics, carreira2018short}. Following \cite{wang2024omnitokenizer}, all video frames are resized to 256 $\times$ 256 resolution for experiments. 
Note that some of the previous works use a resolution of 128 $\times$ 128, which cannot be directly compared to our work due to the difference in task difficulty. However, we still include these results in the table and highlight them in gray.
Moreover, we fine-tune SweetTok’s spatial component on ImageNet \cite{deng2009imagenet} to obtain a strong image tokenizer. The semantic capabilities of SweetTok are tested through few-shot image classification on Real-Name Open-Ended miniImageNet \cite{tsimpoukelli2021multimodal} and few-shot video action recognition on UCF-101, as described in \cite{zhang2020few}. 


\vspace{-3mm}
\paragraph{Evaluation Metrics.} 
For video reconstruction experiments, we evaluate using the Reconstruction Frechet Video Distance (rFVD) \cite{unterthiner2018towards}. For video generation, we use the Generation Frechet Video Distance (gFVD) metric. For image reconstruction, we categorize recent methods by the number of compressed tokens, with each group assessed using the Frechet Inception Distance (FID) \cite{heusel2017gans}.

\begin{figure*}[htbp]
    \centering
    \begin{minipage}[t]{0.33\linewidth}
        \centering
        \includegraphics[width=\linewidth]{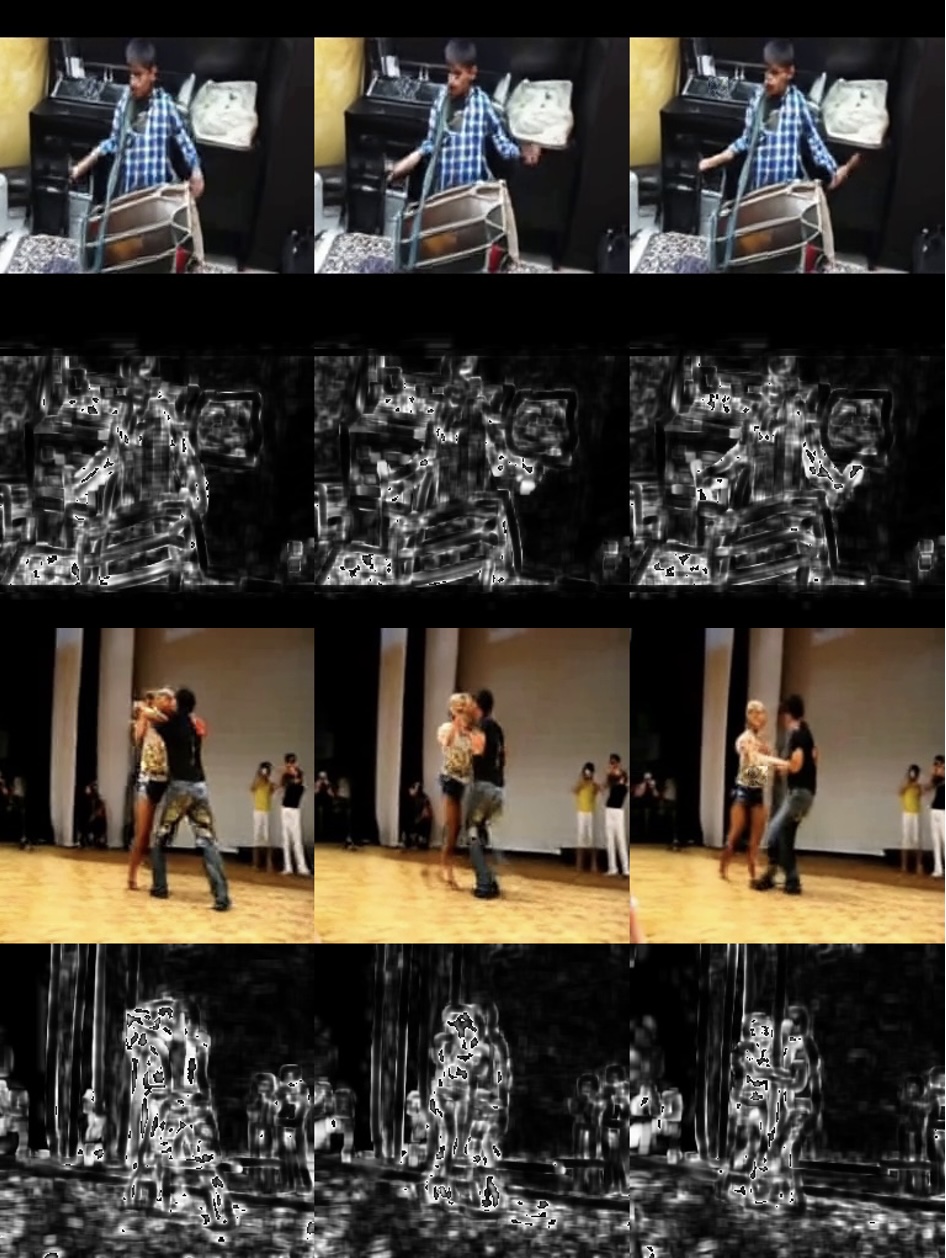} \\
        \vspace{-3mm}
        \caption*{OmniTok~\cite{wang2024omnitokenizer}}
    \end{minipage}
    \begin{minipage}[t]{0.33\linewidth}
        \centering
        \includegraphics[width=\linewidth]{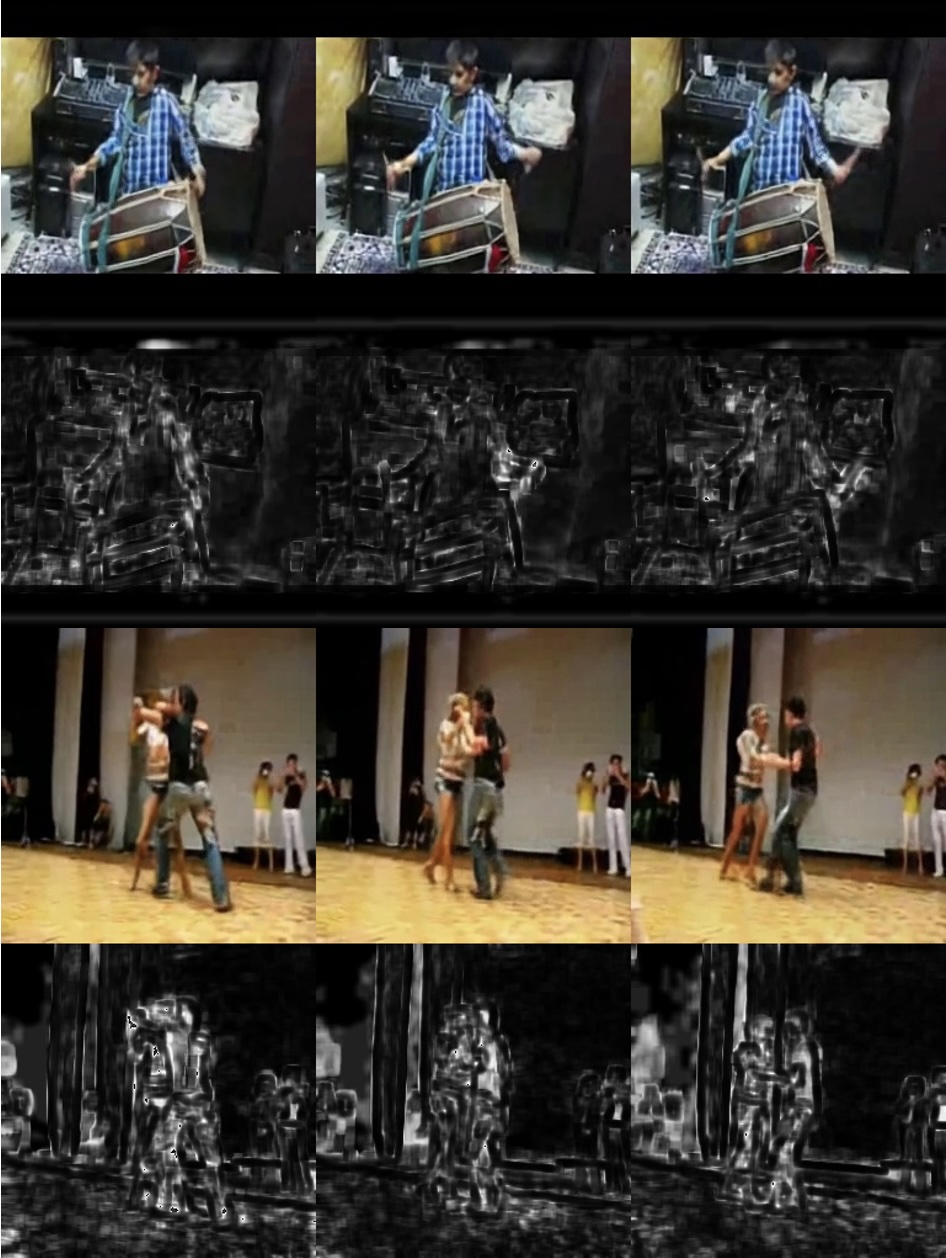} \\
        \vspace{-3mm}
        \caption*{LARP~\cite{wang2024larp}}
    \end{minipage}
    \begin{minipage}[t]{0.33\linewidth}
        \centering
        \includegraphics[width=\linewidth]{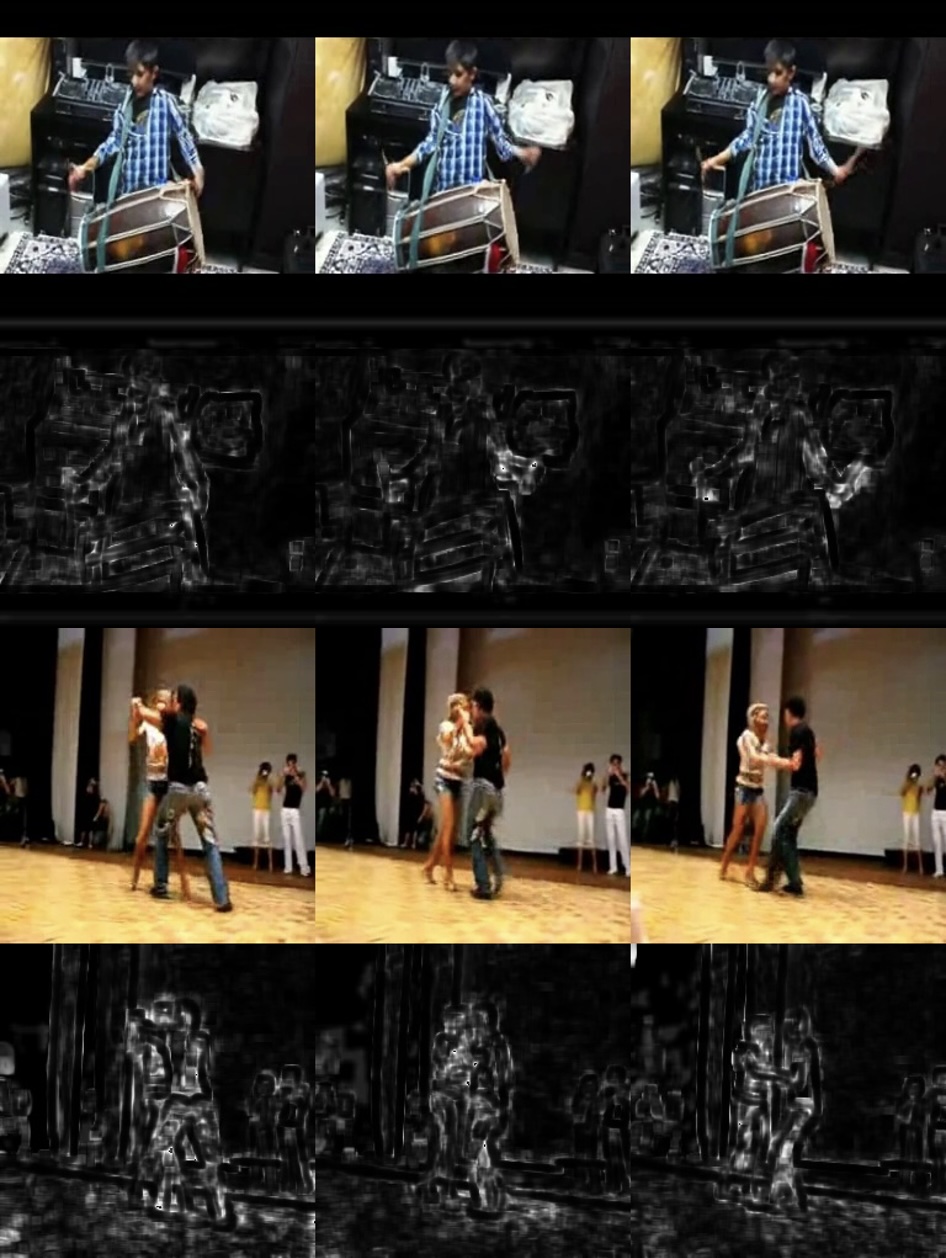} \\
        \vspace{-3mm}
        \caption*{SweetTok (ours)}
    \end{minipage}
    \vspace{-3mm}
\caption{Video reconstruction result on UCF-101 dataset. We also visualize the reconstruction and GT error map, where brighter areas indicate larger errors.
} 
\label{fig:visualization_generation}
\vspace{-3mm}
\end{figure*}

\begin{table}
  \centering
\scalebox{0.85}{
\begin{tabular}{lcc|cc}
\toprule Tokenizer  & \#Tokens & \#Params & \multicolumn{2}{c}{rFVD $\downarrow$}   \\
& &Tokenizer&UCF-101& K-600 \\

\midrule \textcolor{gray}{MAGVIT-V2} \cite{yu2023magvit-v2}  & \textcolor{gray}{1280}& \textcolor{gray}{307M}& \textcolor{gray}{8.6}  &  \textcolor{gray}{-}\\
\textcolor{gray}{LARP-L \cite{wang2024larp}} & \textcolor{gray}{1024}& \textcolor{gray}{193M}  & \textcolor{gray}{20} & \textcolor{gray}{13} \\
\midrule  MaskGIT \cite{chang2022maskgit}   & 4352 & 227M & 240 & 202  \\
VQGAN \cite{esser2020taming}  & 4352& 227M& 299  & 270 \\
TATS \cite{ge2022tats}   & 4096& 32M& 162  & - \\
MAGVIT \cite{yu2023magvit}  & 4096& 158M& 58  & -\\
OmniTok \cite{wang2024omnitokenizer} & 5120& 82.2M  & 42 & 26 \\
LARP-B \cite{wang2024larp}  & 1024&  143M  & 64 & 35 \\
LARP-L \cite{wang2024larp} & 1024 &193M  & 35 & 23 \\
\midrule  
SweetTok$^\ast$  & 5120 & 128M & \textbf{11} & \textbf{8}\\
SweetTok   & 1280 & 128M& 20   & 25 \\
\bottomrule
\end{tabular}
}
\vspace{-2mm}
  \caption{Video reconstruction FVD on the UCF-101 and K-600 dataset, using a frame resolution 256 × 256.    
  ``$\ast$'' denotes training SweetTok without token compression. Lines in ``\textcolor{gray}{gray}'' indicate results at a resolution of 128 $\times$ 128.
  }
  \label{tab:ucf reconstruct}
  \vspace{-2mm}
\end{table}

\paragraph{Implementation Details.} 
SweetTok adopts a spatial-temporal architecture consisting of 8 spatial layers and 4 temporal layers, with both the encoder and decoder configured to a hidden dimension of 512. The latent space dimension is set to 256.
For the LLM codebook quantizer, we exclude words with a frequency below 5, resulting in a selection of 5,078 nouns, 5,403 adjectives, 9,267 verbs, and 1,872 adverbs. This forms a spatial codebook of size 10,481 and a temporal codebook of size 11,139.
The model is trained with a batch size of 8 for 1000K iterations. All training is performed on NVIDIA A100 GPUs.
Adam \cite{kingma2014adam} is employed
for optimization ($\beta_1 = 0.9$ and $\beta_2 = 0.99$). 
During each stage, we use a cosine learning rate scheduler with a max learning rate of 1e-4 and a min learning rate of 1e-5, warmed up by 10K iterations.

\begin{table}
  \centering
  \scalebox{0.8}{
\begin{tabular}{lccc|c}
\toprule Tokenizer & Type & \#Tokens & \#Params& gFVD $\downarrow$ \\ 
&&&Generator&\\
\midrule \textcolor{gray}{MAGVIT} \cite{yu2023magvit} & \textcolor{gray}{AR}& \textcolor{gray}{1024}&\textcolor{gray}{306M}  & \textcolor{gray}{265} \\
\textcolor{gray}{MAGVIT-V2} \cite{yu2023magvit-v2}  & \textcolor{gray}{AR} & \textcolor{gray}{1280} &\textcolor{gray}{307M}& \textcolor{gray}{109} \\
\textcolor{gray}{MAGVIT} \cite{yu2023magvit} & \textcolor{gray}{MLM}  & \textcolor{gray}{1024}&\textcolor{gray}{306M} & \textcolor{gray}{76} \\
\textcolor{gray}{MAGVIT-V2} \cite{yu2023magvit-v2} & \textcolor{gray}{MLM} & \textcolor{gray}{1280}&\textcolor{gray}{307M} & \textcolor{gray}{58} \\
\textcolor{gray}{LARP-L} \cite{wang2024larp}& \textcolor{gray}{AR} & \textcolor{gray}{1024} &\textcolor{gray}{632M} & \textcolor{gray}{57} \\
\midrule CogVideo \cite{hong2022cogvideo} & AR & 6800& 9.4B & 626 \\
TATS \cite{ge2022tats}  & AR& 4096&321M  &332  \\
Video-LaVIT \cite{jin2024videolavit}  & AR& 512& 7B &280 \\
OmniTok \cite{wang2024omnitokenizer}  & AR& 5120 &650M& 191 \\
LARP-L \cite{wang2024larp}& AR & 1024 &632M & 99\\
\midrule 
SweetTok & AR& 1280& 650M  & 84 \\ 
SweetTok & AR& 1280& 1.9B  & \textbf{65} \\ 
\bottomrule
\end{tabular}
}
\vspace{-2mm}
  \caption{
  Class-conditional video generation results on UCF-101. Each video is composed of 17 frames with a resolution of 256 $\times$ 256.
  ``AR'' and ``MLM'' represents autoregressive and masked-language-modeling generator. 
  Lines in ``\textcolor{gray}{gray}'' indicate results at a resolution of 128 $\times$ 128.
  }
  \label{tab:video generation}
  \vspace{-2mm}
\end{table}


\begin{table}
  \centering
\scalebox{0.9}{
\begin{tabular}{lcc|c}
\toprule Tokenizer & \#Tokens & Codebook Size & rFID $\downarrow$ \\
\midrule  VQGAN \cite{esser2020taming} & 256 & 1024 & 7.94 \\
RQ-VAE \cite{lee2022autoregressive} & 256 & 16384 & 3.20 \\
 MaskGIT\cite{yu2023magvit} & 256 & 1024 & 2.28 \\
 LlamaGen-16 \cite{sun2024autoregressive} & 256 & 16384 &2.19\\
 TiTok \cite{titok} & 256 & 4096 & 1.71 \\
 TokenFlow \cite{qu2024tokenflow} & 256& 4096  & 1.03 \\
 \midrule
SweetTok & 256 & 10481 & \textbf{0.73}  \\ \midrule
ViT-VQGAN \cite{yu2021vim} & 1024 & 8192 & 1.28  \\
OmniTok \cite{wang2024omnitokenizer} & 1024 &8192 & 1.11\\
OmniTok$^\diamond$ \cite{wang2024omnitokenizer} & 1024 &8192 & 0.69\\
LlamaGen-8 \cite{sun2024autoregressive} &1024 & 16384 & 0.59\\ \midrule 
SweetTok$^\ast$ & 1024 & 10481 & \textbf{0.37}  \\
\bottomrule
\end{tabular}
}
\vspace{-2mm}
\caption{Image reconstruction FID on the ImageNet dataset, using a resolution of 256 $\times$ 256.
``$\diamond$'' denotes continuous latent space without quantization.
``$\ast$'' denotes training SweetTok without token compression.
}
\label{tab:image reconstruct}
\vspace{-2mm}
\end{table}

\begin{figure*}[htbp]
    \centering
    \begin{minipage}[t]{0.33\linewidth}
        \centering
        \includegraphics[width=\linewidth]{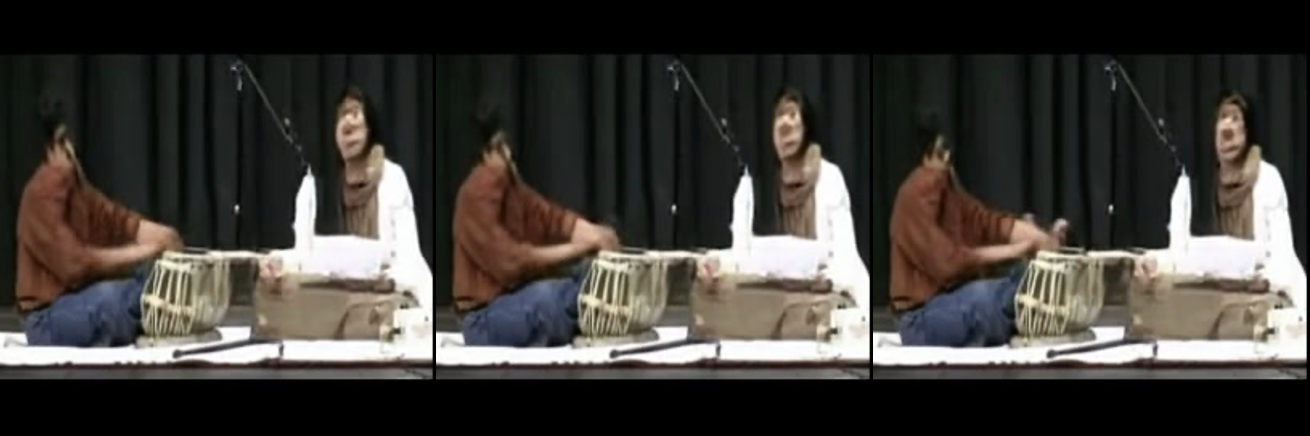} \\
        \includegraphics[width=\linewidth]{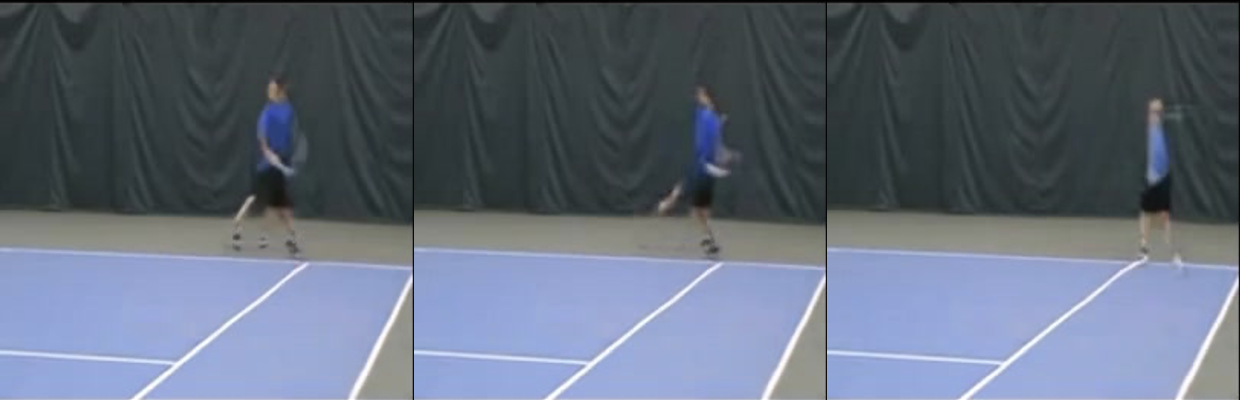} \\
        \includegraphics[width=\linewidth]{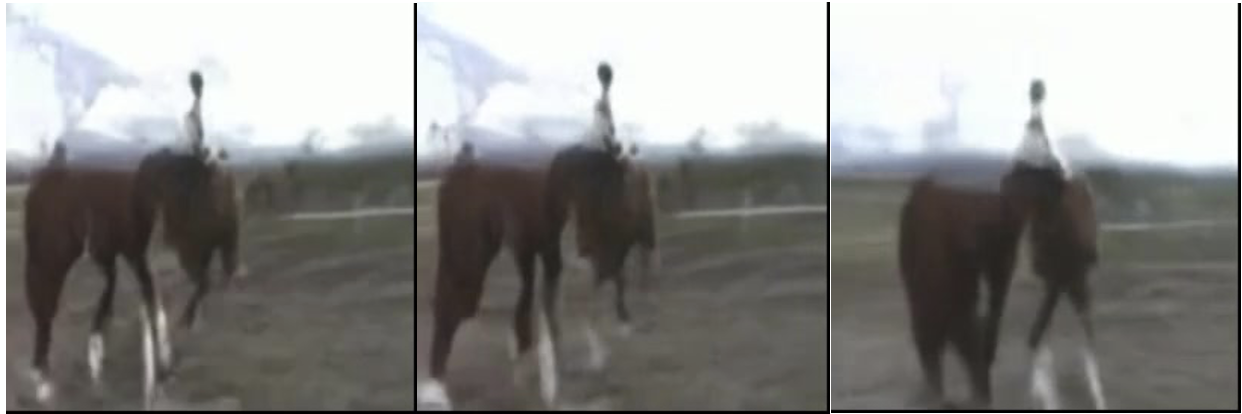} \\
        \vspace{-3mm}
        \caption*{OmniTok~\cite{wang2024omnitokenizer}}
    \end{minipage}
    \begin{minipage}[t]{0.33\linewidth}
        \centering
        \includegraphics[width=\linewidth]{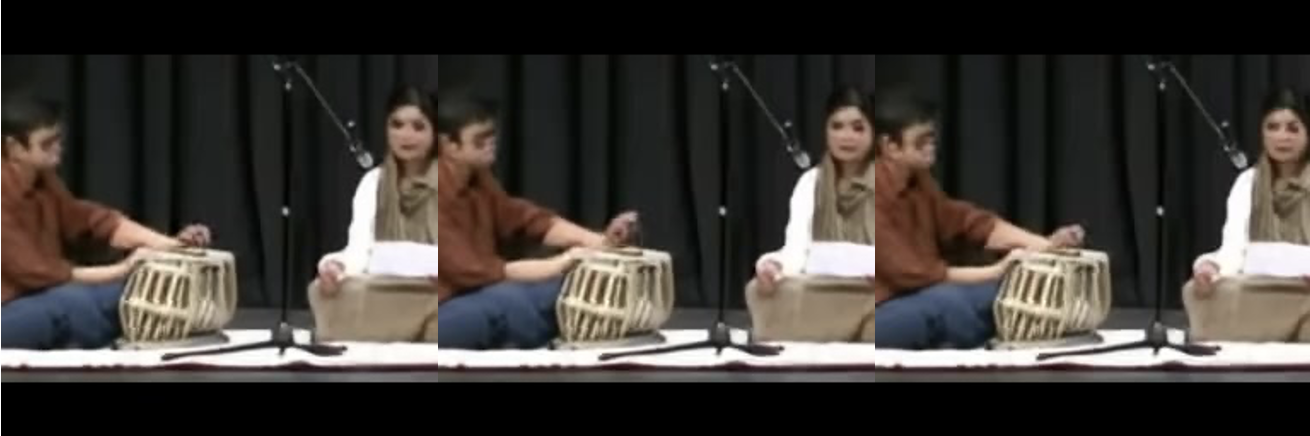}\\
        \includegraphics[width=\linewidth]{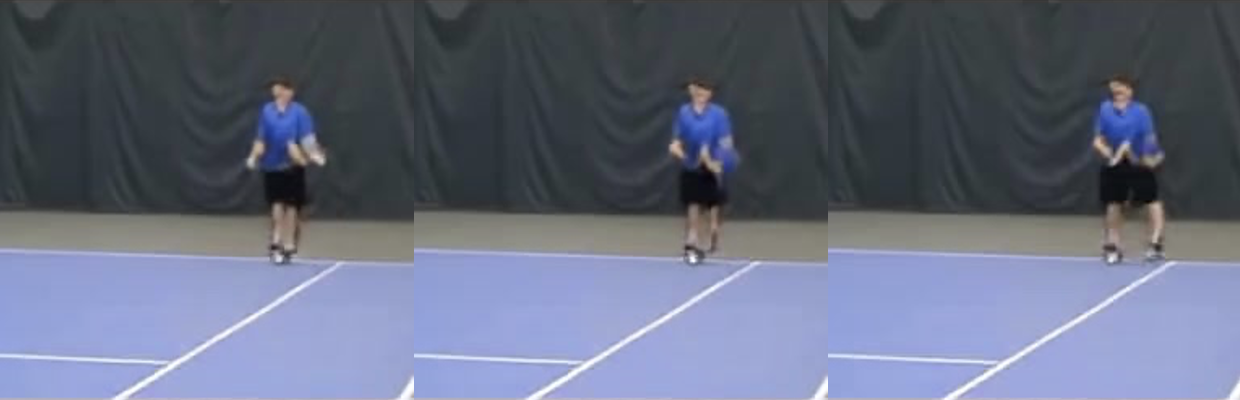}\\
        \includegraphics[width=\linewidth]{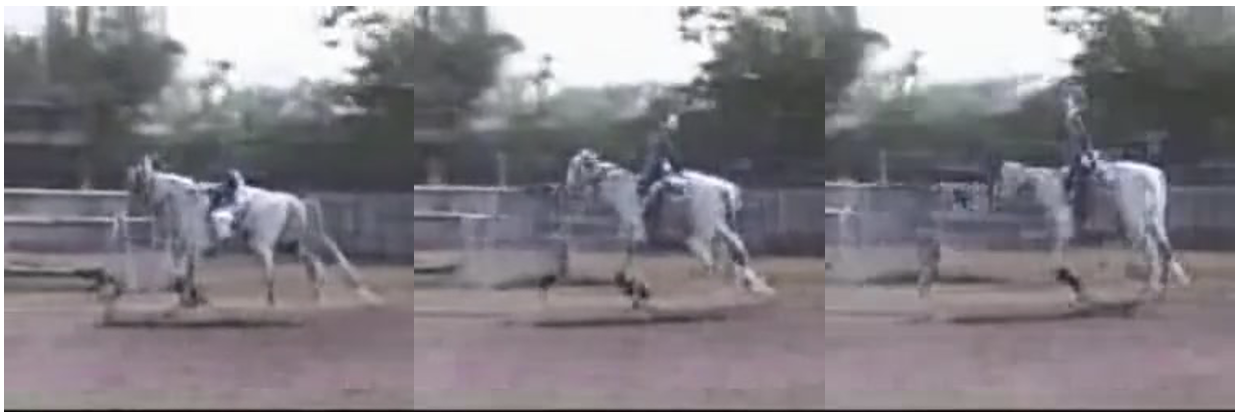} \\
        \vspace{-3mm}
        \caption*{LARP~\cite{wang2024larp}}
    \end{minipage}
    \begin{minipage}[t]{0.33\linewidth}
        \centering
        \includegraphics[width=\linewidth]{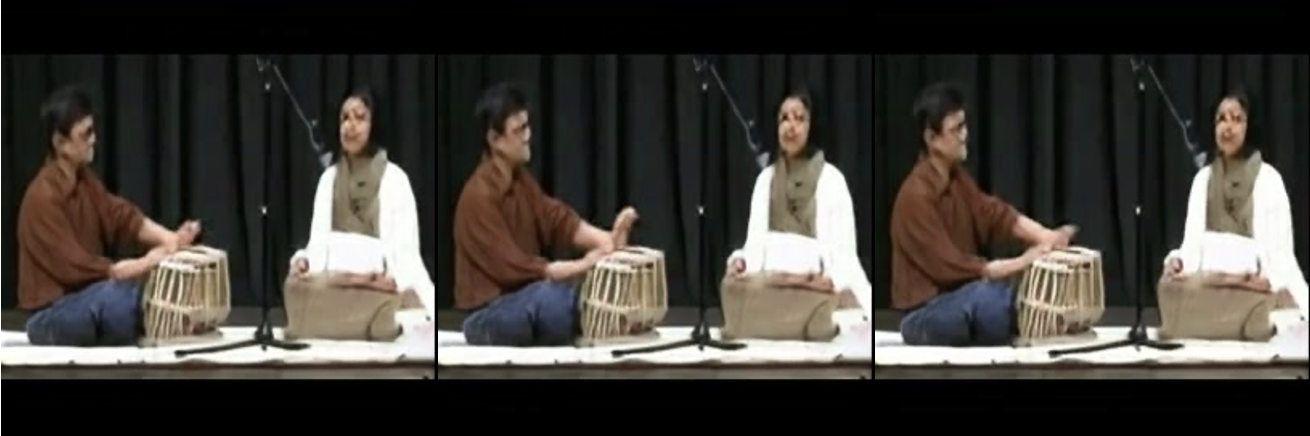}\\
        \includegraphics[width=\linewidth]{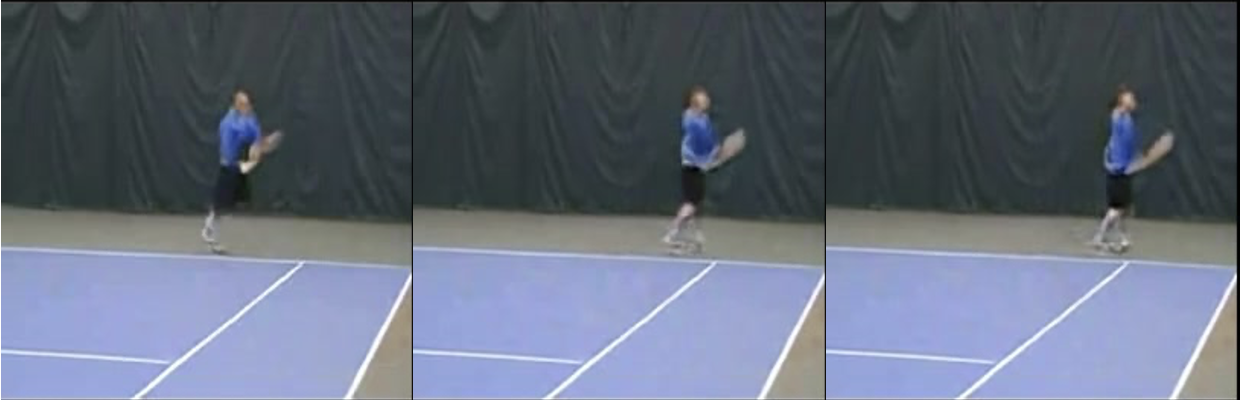}\\
        \includegraphics[width=\linewidth]{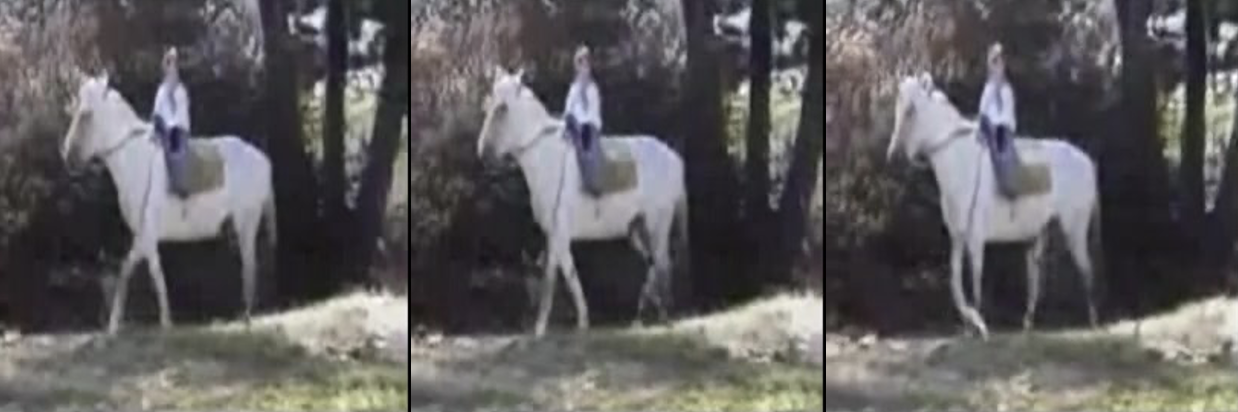} \\
        \vspace{-3mm}
        \caption*{SweetTok (ours)}
    \end{minipage}
    \vspace{-3mm}
\caption{Class-conditional video generation result on UCF-101 dataset. 
The action class label of each row is: 
\textit{``PlayingTabla''}, \textit{``TennisSwing''}, 
and \textit{``HorseRiding''}.
} 
\label{fig:visualization_generation}
\vspace{-3mm}
\end{figure*}

\subsection{Video Reconstruction \& Generation}

We first evaluate the tokenization capability of SweetTok on the UCF-101 and K-600 video datasets. As shown in Table \ref{tab:ucf reconstruct}, SweetTok uses 1,280 tokens (256 spatial tokens and 1,024 temporal tokens), which is four times fewer than OmniTok's 5,120 tokens and comparable with LARP's 1024 tokens. Despite its high compression ratio, our method achieves a 52.3\% improvement on UCF-101 and competitive performance on K-600 compared with OmniTok. With a similar token count, SweetTok, despite its smaller model size compared to LARP-L, delivers a 42.8\% improvement on UCF-101 and comparable results on K-600. Our SweetTok demonstrates significant performance gains, achieving 68.8\% and 28.5\% improvements in rFVD on UCF-101 and K-600 datasets, respectively, compared to LARP-B with similar model size.
Notably, if we increase the token number to 5,120, SweetTok$^\ast$ significantly outperforms all baselines, achieving an rFVD of 10.74 on UCF-101 and 7.51 on K-600.

The generative capability of SweetTok is evaluated on UCF-101 in a class-conditional generation task. SweetTok is used to extract decoupled tokens from UCF videos. These tokens are then concatenated to form training sequences for VideoGPT \cite{yan2021videogpt}, following the same generation protocol as OmniTok.
As shown in Table \ref{tab:video generation}, SweetTok achieves a significant performance improvement, with a gFVD score of 84, 56\% lower than OmniTok's 191 gFVD. This improvement is attributed to SweetTok's effective token compression, which substantially reduces the training complexity for downstream autoregressive models. With equivalent token counts, SweetTok demonstrates superior performance, achieving a 15.1\% improvement in gFVD (84 vs. LARP's 99) at comparable generator sizes. Furthermore, it exhibits scaling law characteristics, with gFVD improving from 84 to 65 as the model scales up to 1.9B parameters.


The visualization results are presented in Figure \ref{fig:visualization_generation}. 
To ensure a fair comparison, we select generated videos with similar appearances, as the generation process inherently involves randomness. The results demonstrate significantly improved detail, such as clearer human facial features and finer table textures. Additionally, SweetTok effectively preserves temporal consistency, even under large motion scenarios. 

\begin{table}
  \centering
  \scalebox{0.9}{
\begin{tabular}{lc|c}
\toprule Compression Method & \#Tokens& rFVD $\downarrow$  \\
\midrule
Vanilla Downsample &  1280&  227.65\\
Vanilla Query-based (LARP~\cite{wang2024larp}) & 1024& 35.15  \\
Decoupled Query-based (DQAE) & 1280 & \textbf{20.46}  \\
\bottomrule
\end{tabular}
}
\vspace{-2mm}
  \caption{Ablation study of different token count compression method for video tokenizers.}
  \label{tab:alblation pts}
  \vspace{-3mm}
\end{table}

\subsection{Image Reconstruction}

\begin{figure}[t]
    \centering
    \begin{minipage}[t]{0.242\linewidth}
        \centering
        \includegraphics[width=\linewidth]{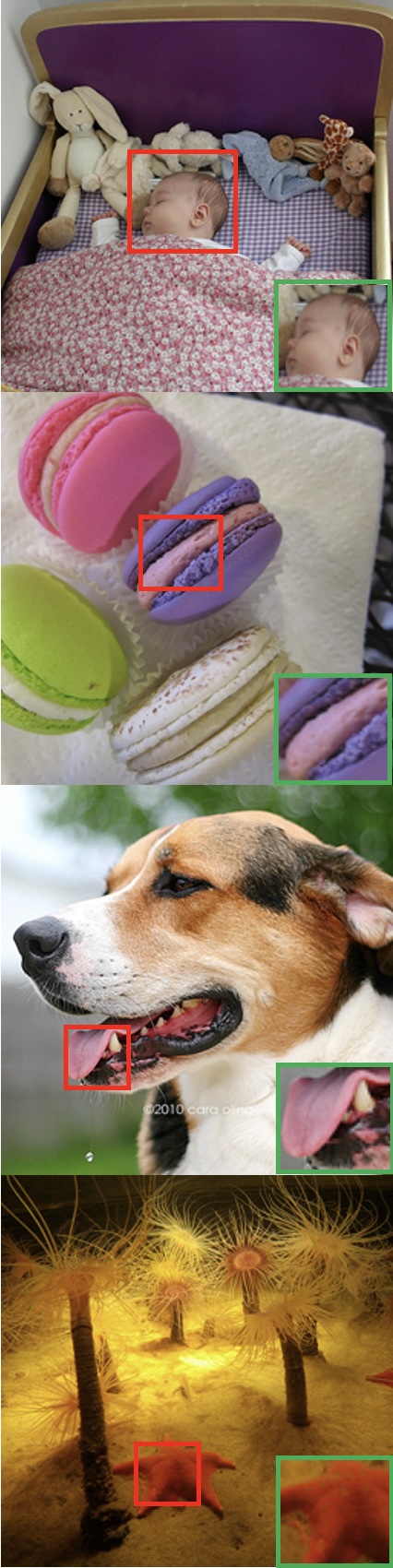} \\
         \vspace{-3mm}
        \caption*{GT}
    \end{minipage}
    \begin{minipage}[t]{0.242\linewidth}
        \centering
        \includegraphics[width=\linewidth]{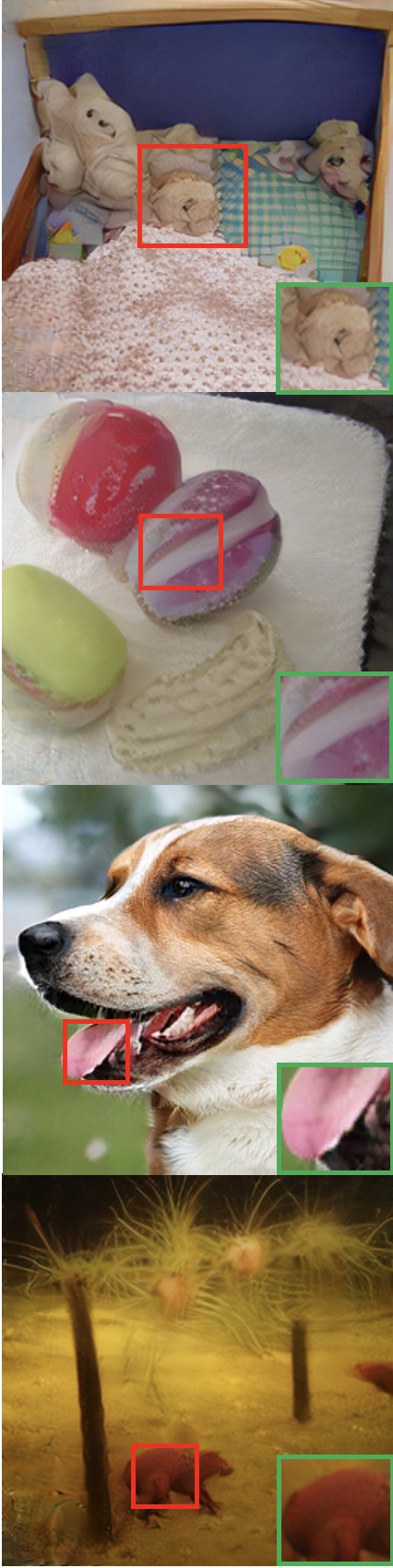}\\
         \vspace{-3mm}
        \caption*{TiTok \cite{titok}} 
    \end{minipage}
        \begin{minipage}[t]{0.242\linewidth}
        \centering
        \includegraphics[width=\linewidth]{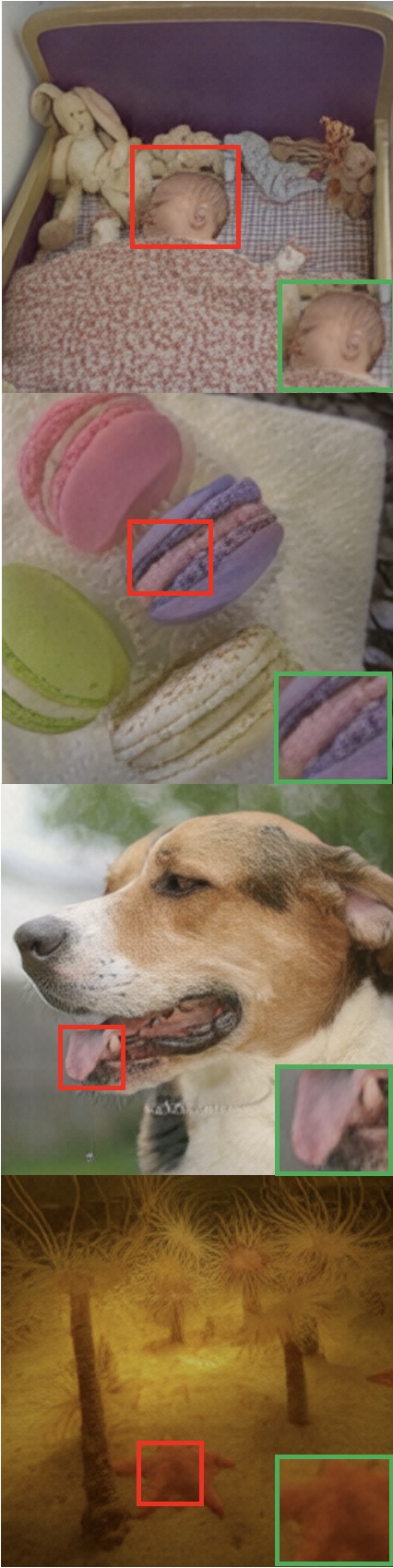}\\
         \vspace{-3mm}
        \caption*{OmniTok \cite{wang2024omnitokenizer}}
    \end{minipage}
        \begin{minipage}[t]{0.242\linewidth}
        \centering
        \includegraphics[width=\linewidth]{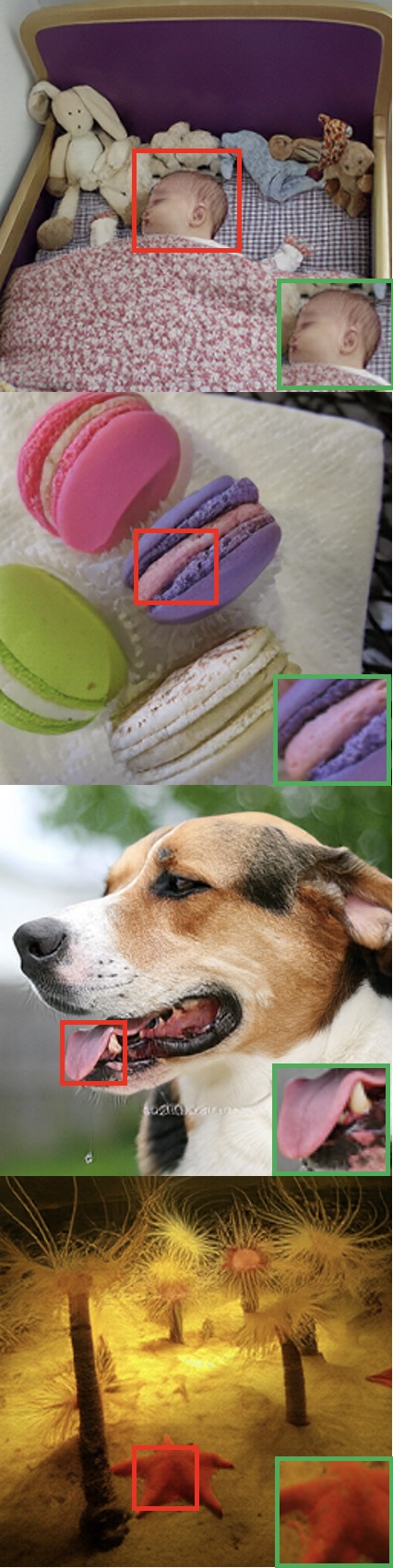}\\
         \vspace{-3mm}
        \caption*{SweetTok} 
    \end{minipage}
    \vspace{-3mm}
\caption{Visualization of image reconstruction results.
} 
\label{fig:img_rec}
\vspace{-3mm}
\end{figure}

To demonstrate the flexibility of our decoupled query design, we show that by directly fine-tuning the spatial branch $DQAE_s$, SweetTok also achieves advanced performance for image reconstruction on ImageNet.
As shown in Table \ref{tab:image reconstruct}, we compare SweetTok with recent methods under various token compression settings.
With 256 spatial tokens, SweetTok outperforms TiTok \cite{titok} by 27.8\%, reducing rFID from 1.01 to 0.73. When using 1,024 spatial tokens, SweetTok$^\ast$ achieves a significant improvement over both VQ-based and non-VQ-based methods (marked $\diamond$), achieving an rFID of 0.37, which surpasses LlammaGen-8 \cite{sun2024autoregressive} by 37.3\%.
Visualization results are shown in Figure~\ref{fig:img_rec}, SweetTok maintains better global appearance and local details.

\subsection{Ablation Studies}

\paragraph{Decoupled Query AutoEncoder.} 
We demonstrate the effectiveness of our spatial-temporal decoupling design for token compression.
As shown in Table \ref{tab:alblation pts}, naively downsampling video tokens by 1D linear-interpolation from 5,120 patches into 1,280 tokens results in poor performance of a rFVD of 227.65. 
Training SweetTok without decoupling (similary to vanilla query-based method in LARP~\cite{wang2024larp}) results in sub-optimal result, obtaining a rFVD of 35.15.
Our proposed decoupling query strategy (DQAE) achieves best result of 20.46 rFVD.
We attribute this improvement to two factors: (1) the flattening operation discards substantial consecutive temporal information, and (2) without decoupling, the model struggles to learn efficiently from the intertwined temporal and spatial information.

\paragraph{Motion-enhanced Language Codebook.}
We evaluate the impact of our architecture design on model performance, focusing on the effects of the motion-enhanced language codebooks (MLC) on UCF-101 datasets. 
As illustrated in Table \ref{tab:alblation arch ucf}, vanilla language codebook design enhances rFVD performance from 29.45 to 24.80, leading to a sub-optimal result.
Notably, our motion-enhanced temporal language codebook significantly benefits video reconstruction tasks, further reducing the rFVD score from 24.80 to 20.46. This underscores the importance of our unique design for video modalities. 
Additionally, we compare different types of language-based embeddings, such as using Qwen-2.5B \cite{bai2023qwen} embeddings in place of CLIP \cite{radford2021clip} embeddings. 
The experiments indicate that naively increasing the complexity of pre-trained language model is not cost-effective for SweetTok.






\subsection{Visual Semantic Comprehension}



\paragraph{Few-Shot Visual Classification.} 


To evaluate the semantic capabilities of SweetTok, we conducted experiments on few-shot image classification and video action recognition tasks. In both experiments, we initially extracted visual tokens using SweetTok and transformed them into natural language words via our language-based codebook. Subsequently, we employed CLIP to compute the similarity between the visual inputs and text embeddings. The top 21 tokens with the highest similarity were selected to form a prompt for prediction using the Qwen LLM.
For the image classification task, we adhered to the V2L protocol \cite{zhu2024v2l-tokenizer}, comparing SweetTok against other language-based visual tokenizers SPAE \cite{yu2024spae} and V2L. In the video action recognition task, we used ARN \cite{zhang2020few} and HF-AR \cite{kumar2021few} as baselines.
The results in Table \ref{tab:classification} indicate that SweetTok achieved an accuracy of 90.8\% on the miniImageNet dataset, surpassing SPAE 89.9\% and V2L 87.6\%. On the UCF-101 dataset, SweetTok attain an average accuracy of 90.1\%, outperforming ARN  83.1\% and HF-AR 86.4\%. These findings demonstrate SweetTok robust semantic understanding and superior performance in both image and video tasks.
We also visualize the effectiveness of our MLC spatial-temporal words for video understanding in the supplementary.



\begin{table}
  \centering
\scalebox{0.92}{
\begin{tabular}{l|c}
\toprule Methods  &  rFVD $\downarrow$ \\
\midrule Baseline (w/o LC) & 29.45  \\ 
  $+$ LC &  24.80\\ 
    $+$ MLC (SweetTok)  & \textbf{20.46} \\ 
\midrule $+$ CLIP \cite{radford2021clip}-based MLC (SweetTok)  &  20.46 \\ 
  $+$ Qwen \cite{bai2023qwen}-based MLC  &  \textbf{20.12}\\ 
\bottomrule
\end{tabular}
}
\vspace{-2mm}
  \caption{
Ablation study of different codebooks, including vanilla language codebook (LC), motion-enhanced language codebook (MLC) and more advanced pre-trained language codebook.
}
  \label{tab:alblation arch ucf}
  \vspace{-3mm}
\end{table}

\begin{table}
  \centering
  \scalebox{0.92}{
\begin{tabular}{l|cccc|c}
\toprule Methods & \multicolumn{3}{c}{ImageNet} &  & UCF-101\\
\midrule
K-way-N-shot & 2-1 & 2-3 & 2-5 & Avg  & 5-5 \\
\midrule 
SPAE \cite{yu2024spae} & 84.8 & 92.5 & 92.6 & 89.9 & - \\
V2L \cite{zhu2024v2l-tokenizer} & 76.3 & 91.2 & \textbf{95.3} & 87.6 & - \\
ARN\cite{zhang2020few} & - & - & - & - & 83.1 \\
HF-AR \cite{kumar2021few} & - & - & - & - & 86.4 \\
SweetTok & \textbf{86.8} & \textbf{90.5} & 95.2 & \textbf{90.8} & \textbf{90.1} \\
\bottomrule
\end{tabular}
}
\vspace{-2mm}
  \caption{Few-shot visual classification accuracy ($\uparrow$), evaluated on both image and video modality.}
  \label{tab:classification}
  \vspace{-3mm}
\end{table}

\label{sec:experiments}

\section{Conclusions}



We present SweetTok, an efficient video tokenization framework that compresses spatial and temporal information through the decoupled query autoencoder. Combined with motion-enhanced language codebook, SweetTok reduces token count for video data more effectively, achieving higher reconstruction fidelity compared to previous state-of-the-arts.
Our approach offers a compact representation of video data, making it well-suited for downstream tasks such as video generation, understanding, marking a significant step in efficient video tokenization.

\label{sec:conclusions}

{
    \small
    \bibliographystyle{ieeenat_fullname}
    \bibliography{main}
}

\clearpage
\setcounter{page}{1}
\maketitlesupplementary

\begin{table}
  \centering
\scalebox{0.95}{
\begin{tabular}{l|c}
\toprule Notations & Explanations   \\
\midrule $DQAE_{s,t}$ & Decoupled query autoencoder \\ 
$\mathcal{P}$ & Patchify Module \\
$\mathcal{E}$ & Encoder\\
$\mathcal{D}$ & Decoder\\
$\mathcal{Q}$ & Quantizer \\
$p_{t,h,w}$ & Downsample ratio \\
$\mathbf{Q}_{s,t,m}$ & Continuous latent query tokens \\
$\mathbf{Z}_{\mathbf{Q}_{s,t,m}}$ & Continuous latent query token \\
$\tilde{\mathbf{Q}}_{s,t}$ & Quantized latent query tokens \\
$\tilde{\mathbf{Z}}_{\mathbf{Q}_{s,t,m}}$ & Quantized latent query token \\
$v_{s,t}$ & Continuous visual feature \\
$\tilde{v}_{s,t}$ & Discrete visual feature \\
$\Delta v$ & Difference of consecutive features \\
$C_{adj, noun, adverb, verb}$ & Codebook text embeddings \\
$c_{adj, noun, adverb, verb}$ & Codebook text embedding \\
$\mathcal{F}$ & Projector network \\
$sg(\cdot)$ & Stop gradient operation \\
\bottomrule
\end{tabular}
}
\vspace{-2mm}
  \caption{Explanations for the notations in the main paper.
  }
  \label{tab:notations}
  \vspace{-2mm}
\end{table}

\section{Experimental Settings}
\subsection{Model Implementation Details}
\paragraph{Visual Tokenizer.} The tokenizer is composed of an encoder $\mathcal{E}$, decoder $\mathcal{D}$, and latent quantizer $\mathcal{Q}$. The tokenizer takes a video clip of 17 consecutive frames with a resolution of 256 $\times$ 256 with the elements normalized to $[-0.5,0.5]$ as input. Then the video clip will be patchified to a resolution of 1 $\times$ 32 $\times$ 32 spatial feature and 4 $\times$ 32 $\times$ 32 temporal feature as illustrated in the main paper. The encoder $\mathcal{E}$ and decoder $\mathcal{D}$ in our tokenizer are both composed of 8 $DQAE_s$ modules and 4 $DQAE_t$ modules with 512 hidden states and 8 attention heads. Each modules consists of self-attention, feed-forward and cross-attention layers. Before the attention computation, the visual features will be reshaped into $[(BT) \times (HW) \times D]$ and $[(BHW) \times T \times D]$ for $DQAE_s$ and $DQAE_t$ modules, respectively. The encoder $\mathcal{E}$ generates 256 spatial and 1024 temporal continuous latent tokens. These tokens are then passed to the quantizer $\mathcal{Q}$, which produces the quantized spatial and temporal latent tokens. The quantizer $\mathcal{Q}$ is composed of a spatial and temporal codebook and a GCN with two hidden layers with hidden dimension of 512 as the projector network $\mathcal{F}$.
To improve the training stability of the visual tokenizer, we adopt exponential moving average (EMA) updates with weight of 0.999 following \cite{titok}.

\paragraph{Language Model.} We utilize VideoGPT following \cite{wang2024omnitokenizer} as the default large language model for the video generative pre-training. All settings follow the protocol of \cite{wang2024omnitokenizer}. 

\begin{table}
  \centering
  \scalebox{0.95}{
\begin{tabular}{lccc}
\toprule Tokenizer & PSNR $\uparrow$ & SSIM $\uparrow$ & LPIPS $\downarrow$ \\ 
\midrule UCF-101&&& \\
\midrule 
LARP \cite{wang2024larp} &27.88&-&0.085 \\
SweetTok  &\textbf{29.27}&\textbf{0.7766}&\textbf{0.070}\\
\midrule ImageNet &&& \\
\midrule 
LlamaGen-16 \cite{sun2024autoregressive} &20.79&0.675& -\\
TokenFlow \cite{qu2024tokenflow} &21.41&0.687& - \\
LlamaGen-8 \cite{sun2024autoregressive} &24.45&0.813&-\\ \midrule 
SweetTok  &\textbf{30.23}&\textbf{0.826}&\textbf{0.068}\\
\bottomrule
\end{tabular}
}
\vspace{-2mm}
  \caption{
  More evaluation results on UCF-101 and ImageNet.
  }
  \label{tab:image generation}
  \vspace{-2mm}
\end{table}


\begin{table*}
  \centering
  \scalebox{0.92}{
\begin{tabular}{lccc}
\toprule Configuration & Language Model & \multicolumn{2}{c}{Tokenizer} \\
&& Image Finetune & Video Training\\
\midrule 
LLM init & VideoGPT& -& -\\ 
Optimizer & AdamW & AdamW & AdamW \\
Optimizer Hyperparameters & $\beta_1 = 0.9$, $\beta_2 = 0.96$ &$\beta_1 = 0.9$, $\beta_2 = 0.99$, $\epsilon = 1e^{-8}$ & $\beta_1 = 0.9$, $\beta_2 = 0.999$\\
Batch size per GPU & 4 & 32  & 12 \\
Peak learning rate & $1e^{-4}$ & $1e^{-4}$  & $1e^{-4}$ \\
Discriminator peak learning rate & - & - & $1e^{-4}$ \\
Learning rate schedule & Cosine & Cosine & Cosine\\
Training steps & 1000K & 500K & 1000K \\
Discriminator start steps & - & -  & 20K \\
Warm-up steps & 10K & 10K  & 10K\\
Weight decay & 0.03 & $1e^{-4}$  & $1e^{-4}$\\
Numerical precision & float16 & float16 & bfloat16 \\
\bottomrule
\end{tabular}
}
\vspace{-3mm}
  \caption{The detailed training hyperparameters of SweetTok.}
  \label{tab:training settings}
  \vspace{-3mm}
\end{table*}

\subsection{Training Datasets}
\paragraph{UCF-101.} UCF-101 is a large-scale action recognition dataset consisting of 13,320 videos with 9537 for training and 3783 for testing across 101 action categories. The dataset includes videos with significant variations in camera motion, object appearance, scale, viewpoint, cluttered backgrounds, and lighting conditions, making it one of the most challenging datasets for action recognition. 

\paragraph{Kinetic-600.} Kinetics-600 is a large-scale action recognition dataset containing approximately 480K videos across 600 action categories. The dataset is split into 390K training, 30K validation, and 60K test videos. Each video is a 10-second clip extracted from raw YouTube footage, focusing on key action moments.  

\paragraph{ImageNet-1K.} ImageNet-1K is a widely used subset of the larger ImageNet dataset, specifically designed for image classification tasks. It contains 1.2 million labeled images across 1,000 distinct categories, ranging from animals and plants to everyday objects and scenes. Each category in ImageNet-1K includes a set of training images, along with separate validation and test sets for model evaluation. The dataset is widely used for researchs in computer vision.

\subsection{Notations}
The meaning of our notations appeared in the main paper are explained in Table \ref{tab:notations}.

\subsection{Training Settings}
The detailed training hyper-parameter settings for SweetTok are
reported in Table \ref{tab:training settings}. During video training, we train the first 500K steps using proxy code following \cite{titok} to accelerate training. The remaining 500K is trained without proxy code.

\section{Additional Results}

\subsection{More evaluation metrics}
We assess SweetTok using additional metrics: PSNR, SSIM, and LPIPS. As shown in Table \ref{tab:image generation}, SweetTok outperforms all baselines on both video and image datasets, further validating the superiority of our model design.


\subsection{More Visualizations}
\begin{figure*}[htbp]
    \centering
    \includegraphics[width=\linewidth]{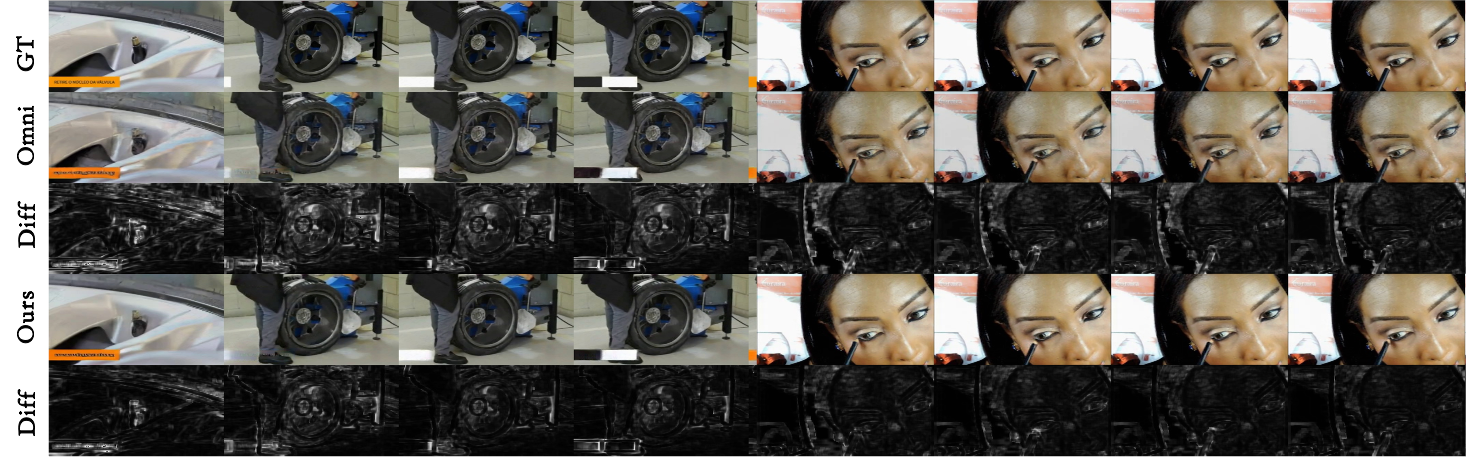}
    \includegraphics[width=\linewidth]{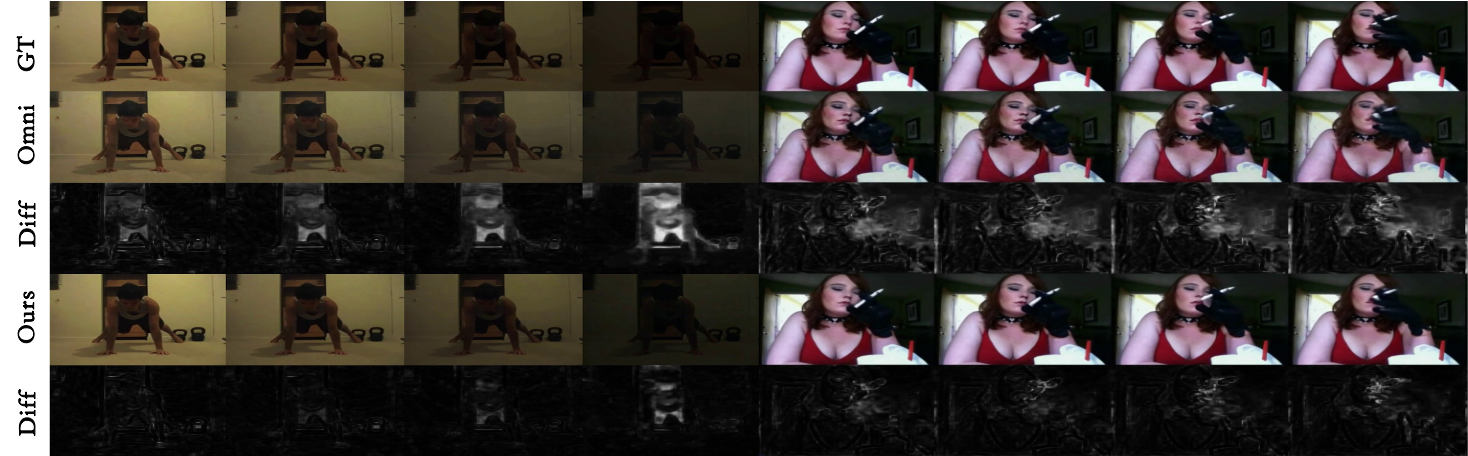}
    \includegraphics[width=\linewidth]{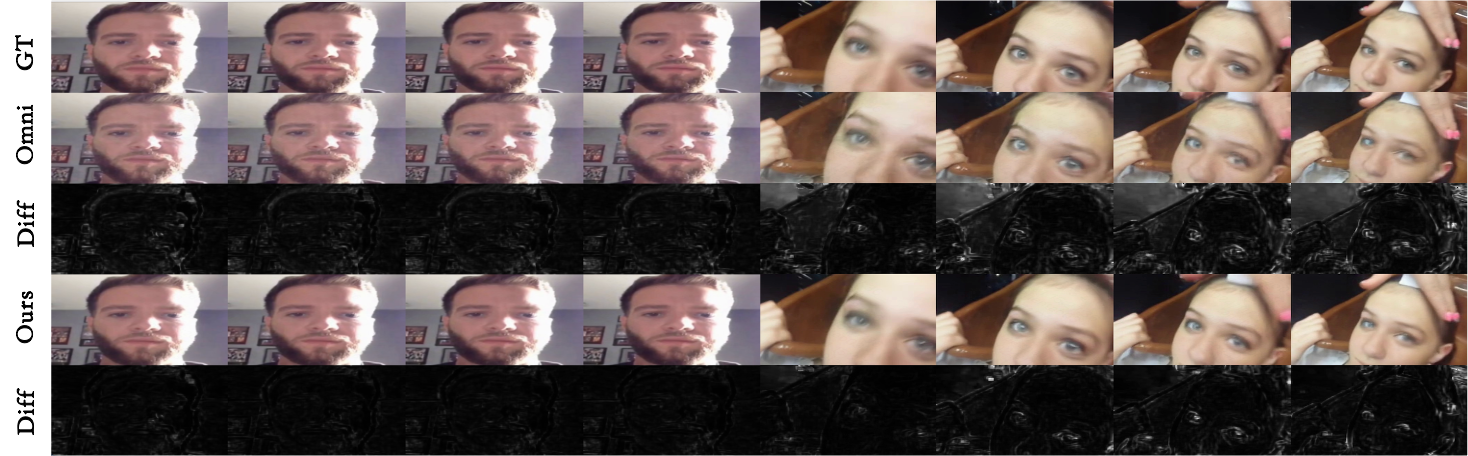}
    \includegraphics[width=\linewidth]{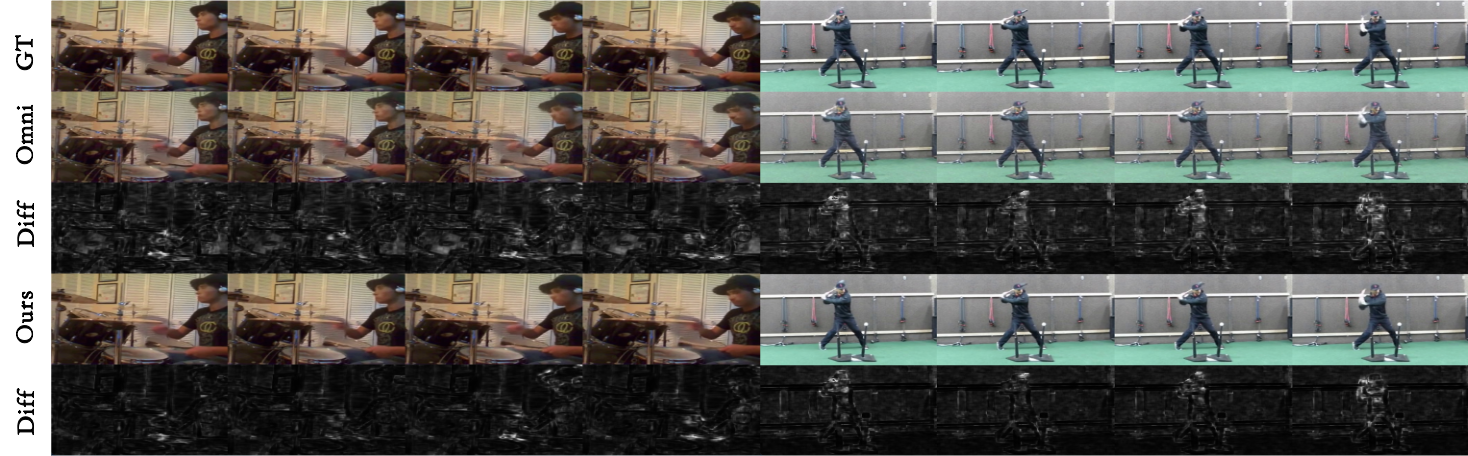}
\caption{Comparison of the reconstruction results of OmniTokenizer and SweetTok on K-600 dataset, where "Diff" represents the pixel difference between the ground truth and the models.} 
\label{fig:visualization_K600_reconstruction}
\vspace{-3mm}
\end{figure*}

\begin{figure*}[htbp]
    \centering
    \includegraphics[width=\linewidth]{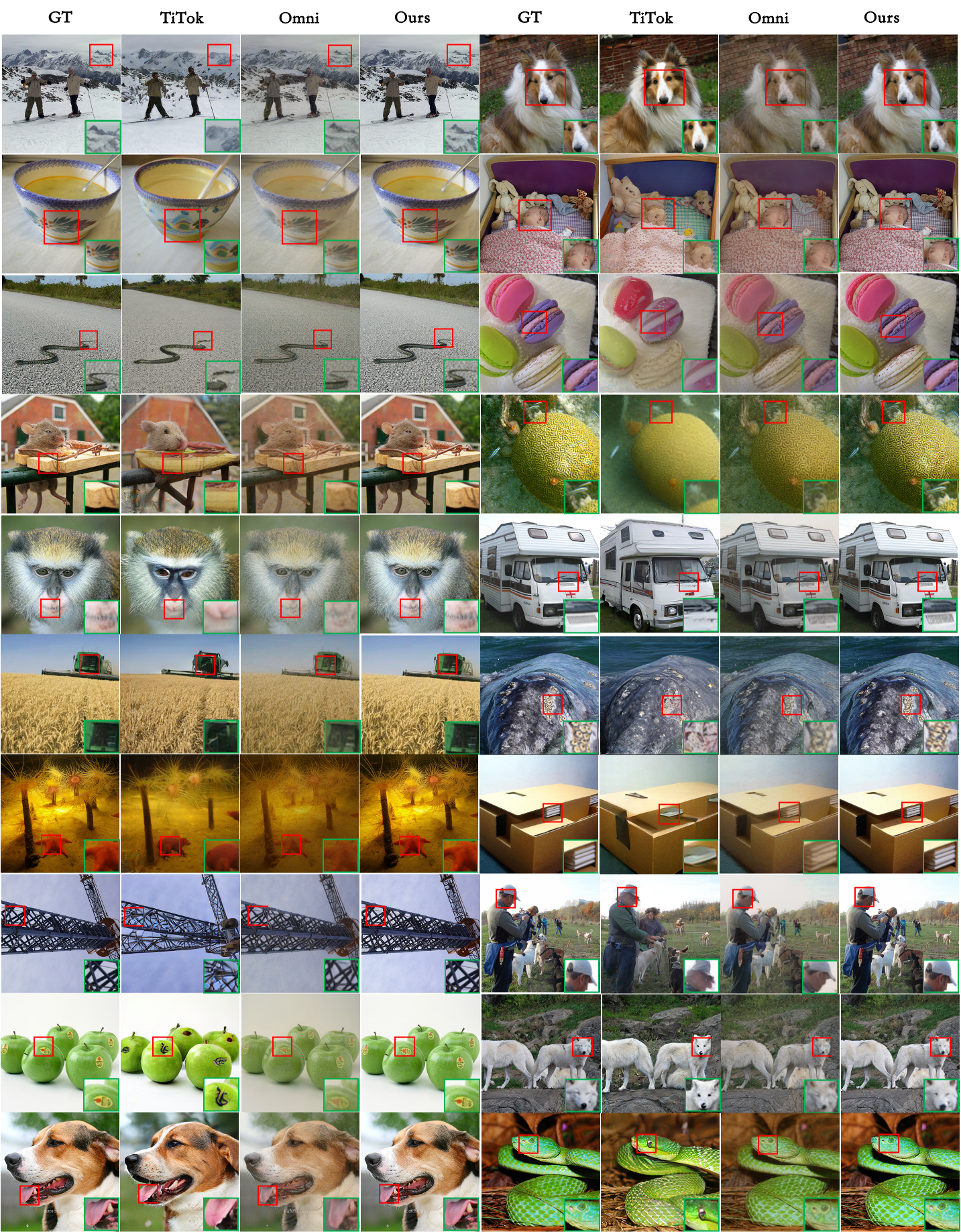}
\caption{Comparison of the reconstruction results of TiTok, OmniTokenizer, and SweetTok on ImageNet-1K dataset. Differences are selected by the red blocks and highlighted in the grean blocks.} 
\label{fig:visualization_img_reconstruction}
\vspace{-3mm}
\end{figure*}

Fig \ref{fig:visualization_img_reconstruction} and Fig \ref{fig:visualization_K600_reconstruction} visualize the reconstruction results for the UCF-101 and K-600 datasets. The pixel-level differences between ground truth and model are shown, with brighter areas indicating greater disparity and darker areas reflecting consistency. As shown, SweetTok exhibits fewer reconstruction differences compared to OmniTokenizer, demonstrating its superior performance.

Fig \ref{fig:visualization_img_reconstruction}  visualize the reconstruction results of SweetTok on ImageNet-1K. For reconstruction, differences between models are highlighted in red blocks, with details shown in green blocks. Clearly, SweetTok outperforms all baselines by a significant margin.

\begin{figure*}[htbp]
    \centering
    \includegraphics[width=\linewidth]{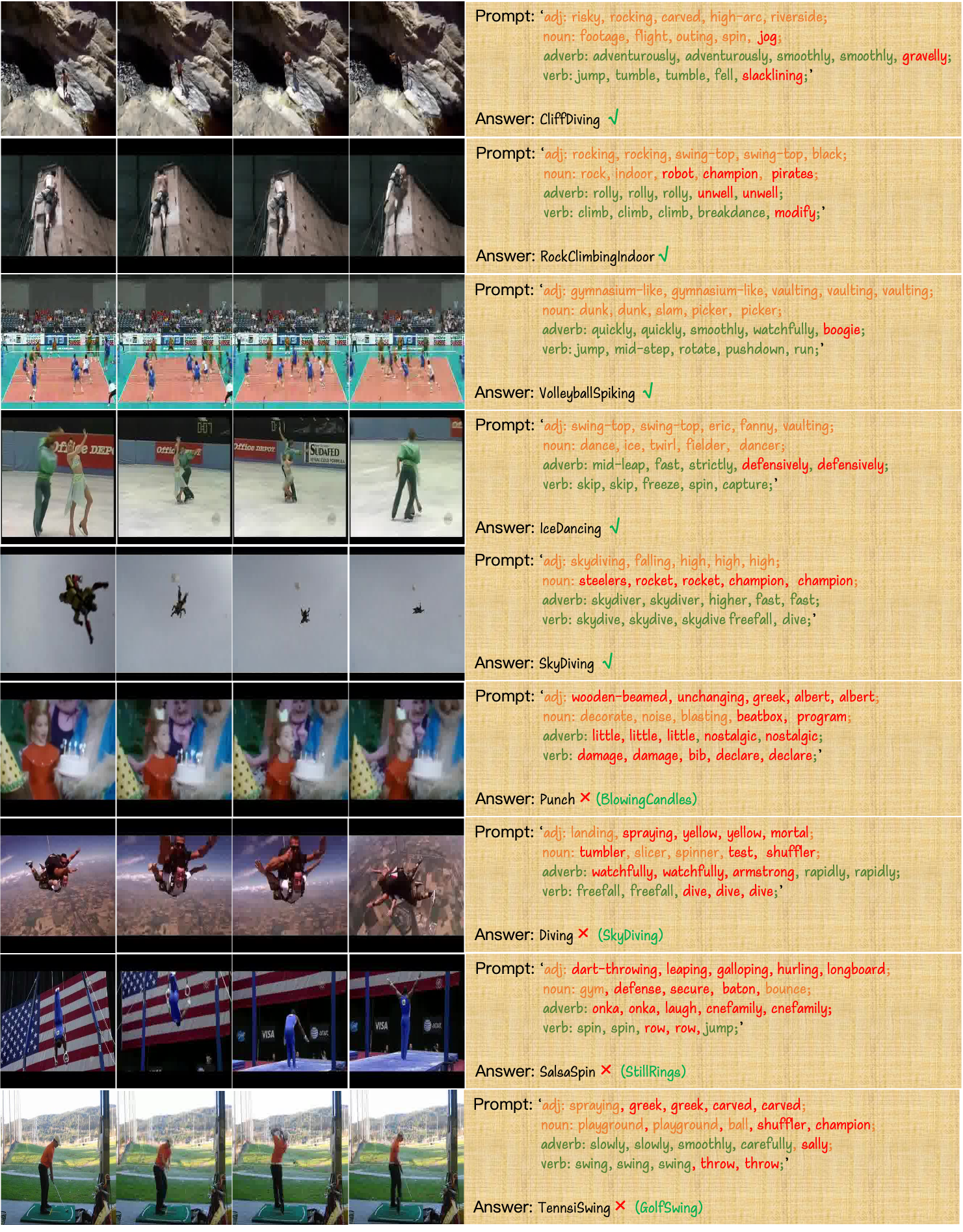}
\caption{Semantic words visualization for UCF-101. The visualization is based on few shot video action recognition tasks. 
} 
\label{fig:semantic}
\vspace{-3mm}
\end{figure*}

Finally, we visualize the words from our MLC in Fig \ref{fig:semantic}, based on few-shot video action recognition tasks on the UCF-101 dataset. We use adjectives, nouns, adverbs, and verbs as prompts to Qwen LLM for action prediction. Green and orange indicate meaningful words, while red marks meaningless ones. The visualization shows that correct verb words consistently lead to accurate predictions, even when other words are irrelevant, highlighting the importance of our MLC modules for video action recognition.


\begin{figure*}[htbp]
    \centering
    \includegraphics[width=\linewidth]{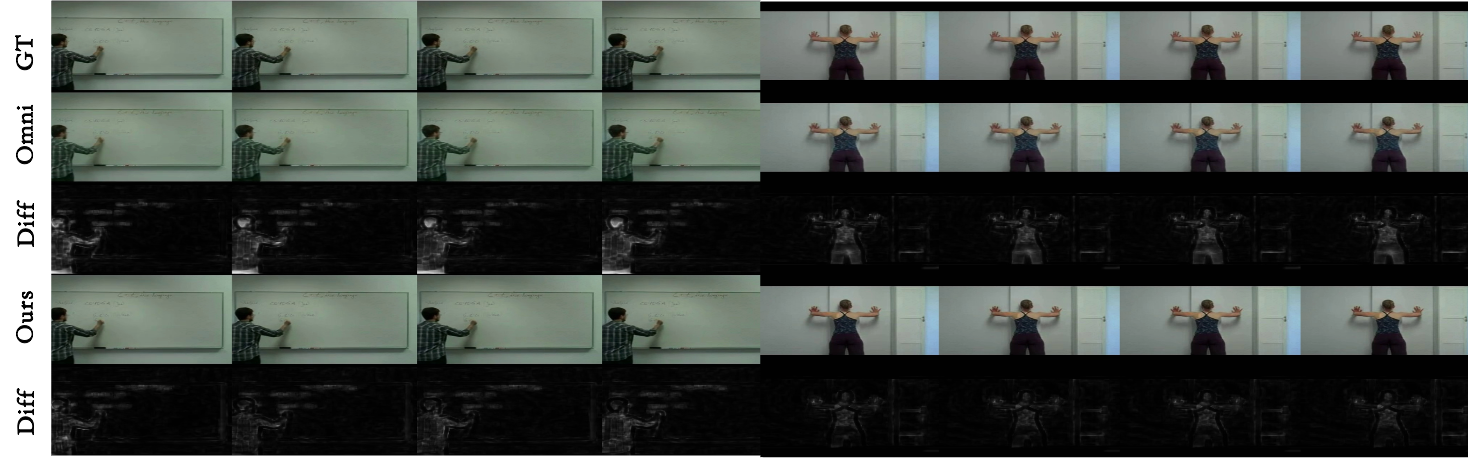}
    \includegraphics[width=\linewidth]{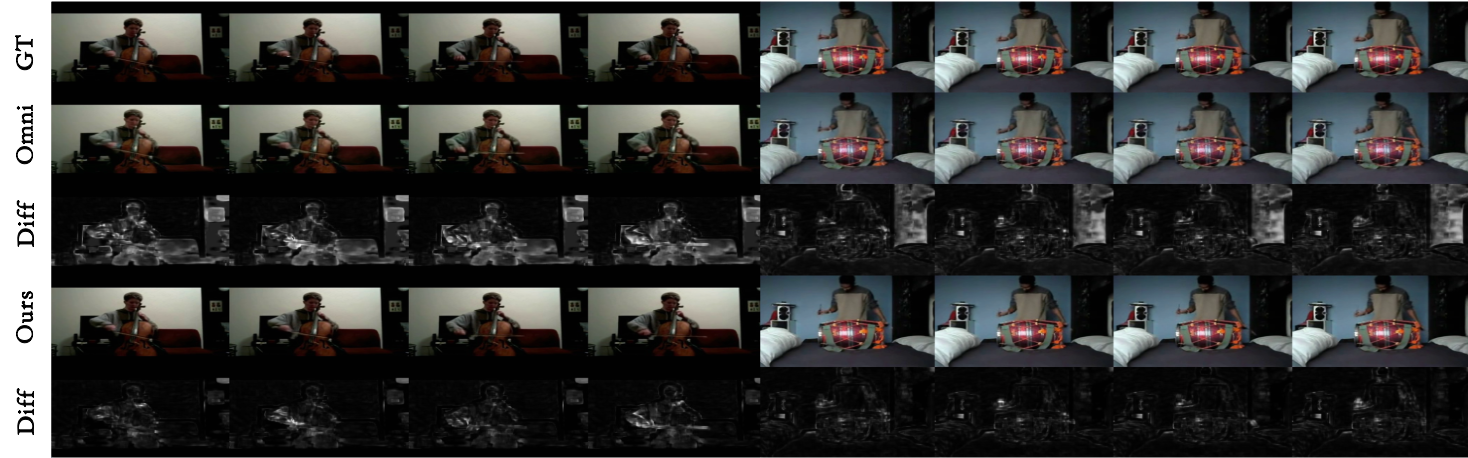}
    \includegraphics[width=\linewidth]{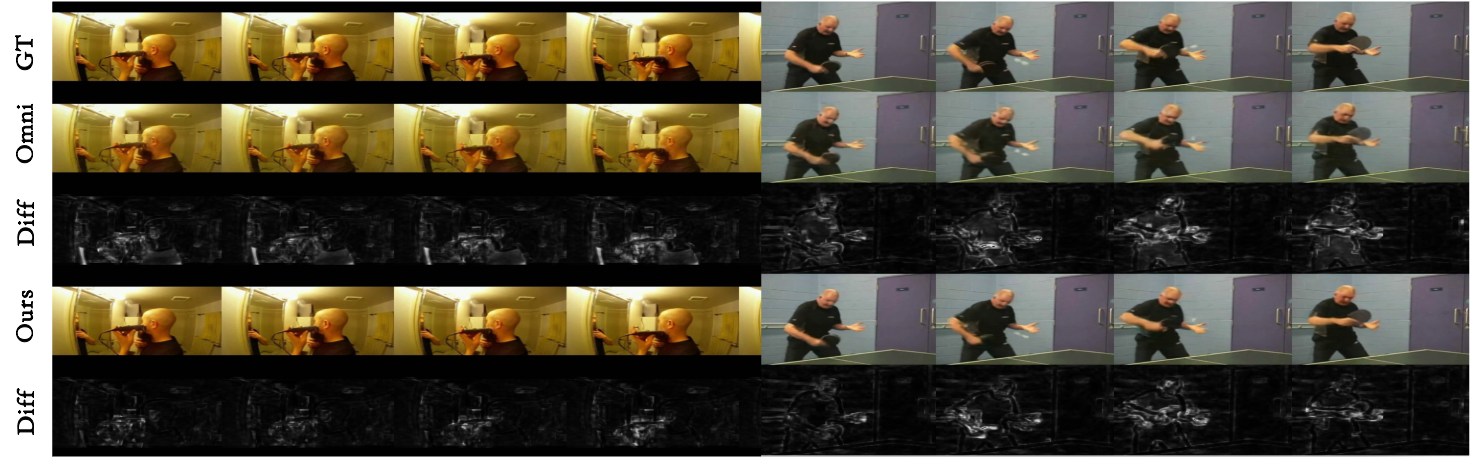}
    \includegraphics[width=\linewidth]{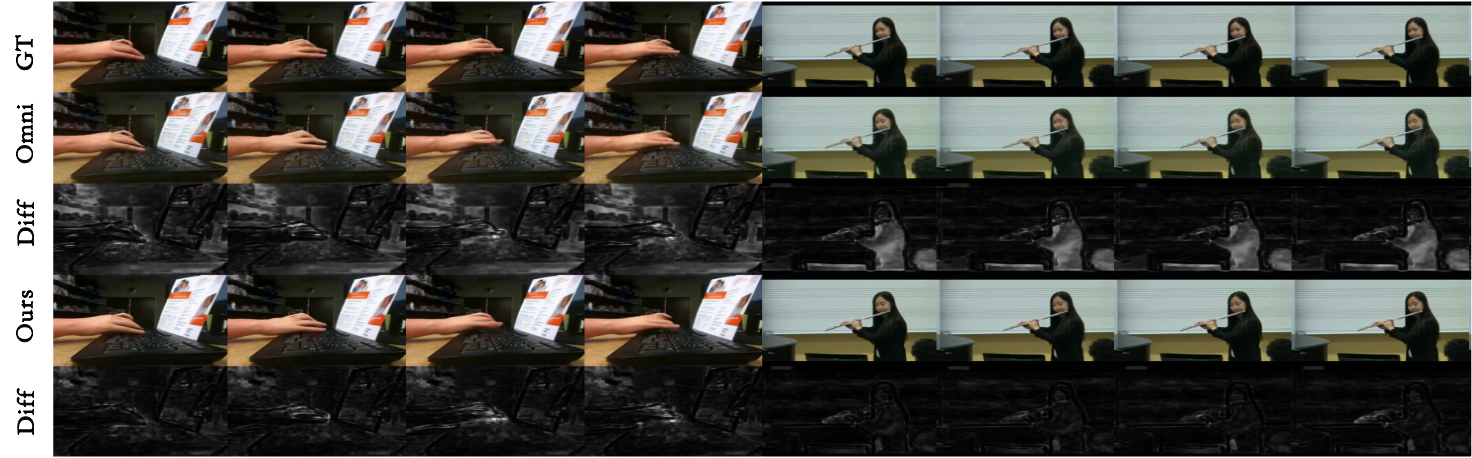}
\caption{Comparison of the reconstruction results of OmniTokenizer and SweetTok on UCF-101 dataset, where "Diff" represents the pixel difference between the ground truth and the models.} 
\label{fig:visualization_ucf_reconstruction}
\vspace{-3mm}
\end{figure*}

\section{Limitations}
Our tokenizer is not suitable for tasks requiring precise semantic understanding, like VQA, because the MLC is trained in an unsupervised manner. Without additional constraints, such as contrastive learning between image features from Qwen-VLM and text embeddings in our codebook, aligning the image and text domains is challenging. A promising direction for future work is to enhance SweetTok into a semantically strong tokenizer by contrastive learning.


\end{document}